\documentclass{article} 
\usepackage{iclr2021_conference,times}
\usepackage{graphicx} 
\usepackage{subcaption}

\usepackage{tikz}

\usepackage{amsmath,amsfonts,bm}

\def\eqref#1{equation~\ref{#1}}

\def\1{\bm{1}}

\DeclareMathAlphabet{\mathsfit}{\encodingdefault}{\sfdefault}{m}{sl}
\SetMathAlphabet{\mathsfit}{bold}{\encodingdefault}{\sfdefault}{bx}{n}

\usepackage{hyperref}
\usepackage{url}
\usepackage{caption}
\usepackage{subcaption}
\usepackage{multirow}
\usepackage{chngpage}
\usepackage{tabularx}

\usepackage{wrapfig}
\usepackage{enumitem}
\usepackage[utf8]{inputenc}
\usepackage{tikz}
\usetikzlibrary{matrix,shapes,arrows,positioning,chains}

\newcommand*\samethanks[1][\value{footnote}]{\footnotemark[#1]}

\title{Are all outliers alike? \\ A Taxonomy of Out of Distribution Samples}

\title{Are all outliers alike? \\ On Understanding the Diversity of Outliers\\ for Detecting OODs}

\author{Ramneet Kaur\thanks{Ramneet Kaur is a graduate student at the Department of Computer and Information Science, University of Pennsylvania, Philadelphia, USA. This work was done when she was a summer intern at SRI.}\ , \; Susmit Jha\thanks{Computer Science Laboratory, SRI International, Menlo Park, CA 94025, USA} \ , Anirban Roy\samethanks\ , Oleg Sokolsky\thanks{Department of Computer and Information Science, University of Pennsylvania, Philadelphia, USA.} \ \& Insup Lee\samethanks  \\

\texttt{ramneetk@seas.upenn.edu,\{susmit.jha,anirban.roy\}@sri.com,} \\
\texttt{\{sokolsky, lee\}@cis.upenn.edu}
}

\newcommand{\noop}[1]{{}}

\newcommand{\RK}[1]{{\color{red}{#1}}}

\newcommand{\AR}[1]{{\color{magenta}{#1}}}

\newcommand{\newinp}{\hat x}
\newcommand{\newclass}{\hat c}

\tikzset{
decision/.style={
    diamond,
    draw,
    text width=4em,
    text badly centered,
    inner sep=0pt
},
block/.style={
    rectangle,
    draw,
    text width=10em,
    text centered,
    rounded corners
},
sblock/.style={
    rectangle,
    draw,
    text width=6em,
    text centered,
    rounded corners
},
cloud/.style={
    draw,
    ellipse,
    minimum height=2em
},
descr/.style={
    fill=white,
    inner sep=2.5pt
},
connector/.style={
    -latex,
    font=\scriptsize
},
rectangle connector/.style={
    connector,
    to path={(\tikztostart) -- ++(#1,0pt) \tikztonodes |- (\tikztotarget) },
    pos=0.5
},
rectangle connector/.default=-2cm,
straight connector/.style={
    connector,
    to path=--(\tikztotarget) \tikztonodes
}
}

\iclrfinalcopy 
\begin{document}

\maketitle

\begin{abstract}
Deep neural networks (DNNs) are known to produce incorrect predictions with very high confidence on out-of-distribution (OOD) inputs. This limitation is one of the key challenges in the adoption of deep learning models in high-assurance systems such as autonomous driving, air traffic management, and medical diagnosis. This challenge has received significant attention recently, and several techniques have been developed to detect inputs where the model's prediction cannot be trusted. These techniques use different statistical, geometric, or topological signatures. This paper presents a taxonomy of OOD outlier inputs based on their source and nature of uncertainty. We demonstrate how different existing detection approaches fail to detect certain types of outliers. We utilize these insights to develop a novel integrated detection approach that uses multiple attributes corresponding to different types of outliers. Our results include experiments on CIFAR10, SVHN and MNIST as in-distribution data and Imagenet, LSUN, SVHN (for CIFAR10), CIFAR10 (for SVHN), KMNIST, and F-MNIST as OOD data across different DNN architectures such as ResNet34, WideResNet, DenseNet, and LeNet5.
\noop{Our results 
include improvement in the true negative rate, i.e., the fraction of detected out of distribution samples (SVHN), from 53.16\% (by current state of the art) to 88.2\% on ResNet when 95\% of in-distribution (CIFAR-10) samples are correctly detected.}

\end{abstract}

\section{Introduction}
\label{sec:intro}


Deep neural networks (DNNs) have achieved remarkable  performance-levels in many areas such as computer vision~\citep{img-classification}, speech recognition~\citep{speech-recog}, and text analysis~\citep{text-analysis}. But their deployment in the safety-critical systems 
such as self-driving vehicles~\citep{ML-App7}, aircraft collision avoidance~\citep{NNinAircraft}, and medical diagnoses~\citep{NNclinically} is hindered by their brittleness. 
One major challenge is the inability of DNNs to be self-aware of when new inputs are outside the training distribution and likely to produce incorrect predictions. It has been widely reported in literature~\citep{guo2017calibration,baseline} that deep neural networks 
exhibit overconfident incorrect predictions on inputs which are outside the training distribution. 
 The responsible deployment of deep neural network models in high-assurance applications necessitates detection of  out-of-distribution (OOD) data so that DNNs can abstain from making decisions on those. 



Recent approaches for OOD detection consider different statistical, geometric or topological  signatures in data that  
differentiate OODs from the training distribution. For example, the changes in the softmax scores due to input perturbations and temperature scaling have been used to detect OODs~\citep{baseline, odin, temp-scaling}. \citet{dknn} use the conformance among the labels of the nearest neighbors while \citet{contrastive} use cosine similarity (modulated by the norm of the feature vector) to the nearest training sample for the detection of OODs. \citet{mahalanobis} consider the Mahalanobis distance of an input from the in-distribution data to detect OODs. Several other metrics such as reconstruction error~\citep{vae-recon-err}, likelihood-ratio between the in-distribution and OOD samples~\citep{likelihood-ratio}, trust scores (ratio of the distance to the nearest class different from the predicted class and the distance to the predicted class)~\citep{trust-score}, density function~\citep{energy, oe}, probability distribution of the softmax scores~\citep{kl-div,self-supervised,contrastive,oe} have also been used to detect OODs. All these methods attempt to develop a uniform approach with a single signature to detect all OODs accompanied by empirical evaluations that use datasets such as CIFAR10 as in-distribution data and other datasets such as SVHN as OOD. 

Our study shows that OODs  can be of diverse types with different defining characteristics. Consequently, an integrated approach that takes into account the diversity of these outliers is needed for effective OOD detection.
We make the following three contributions in this paper: 
\begin{itemize}[topsep=0pt, leftmargin=*]
    \item \textbf{Taxonomy of OODs.}  We define a taxonomy of OOD samples that classify OODs into different types based on aleatoric vs epistemic uncertainty~\citep{ML-uncertainty-survey}, distance from the predicted class vs the distance from the tied training distribution, and uncertainty in the principal components vs uncertainty in  non-principal components with low variance. 
    \item \textbf{Incompleteness of existing uniform OOD detection approaches.}  We examine the limitations of the state-of-the-art approaches to detect various types of OOD samples. We observe that not all outliers are alike and existing approaches fail to detect particular types of OODs. We use a toy dataset comprising two halfmoons as two different classes to demonstrate these limitations. 
    \item \textbf{An integrated OOD detection approach.} We  propose an integrated approach that can detect different types of OOD inputs. We demonstrate the effectiveness of our approach on several benchmarks,
     and compare against state-of-the-art OOD detection approaches such as the ODIN~\citep{odin} and Mahalanobis distance method~\citep{mahalanobis}.
\end{itemize}

\noop{\AR{We argue that all ODDs samples are not alike. The above-mentioned approaches aim to detect the ODD samples based on one or more properties of them. However, one single approach cannot detect all types of OOD samples. Motivated by this observation, we first create a taxonomy of OOD samples and then propose a unified approach to detect various OOD samples. Following~\cite{ML-uncertainty-survey}, we classify OOD samples in two main categories based on the associated source of uncertainties: 1) Aleatoric OOD samples where the sample are likely to overlap with multiple in-distribution classes and 2) Epistemic OOD samples where the sample are far from all the in-distribution classes (Fig.~\ref{fig:OOD_types},~\ref{fig:OOD_flow}). Our three central contributions are: 
\begin{itemize}[noitemsep, topsep=0pt, leftmargin=*]
    \item We define a taxonomy of OOD samples and examine the limitations of the state-of-the-art approaches to detect various types of OOD samples. 
    \item We propose a unified approach that can detect a diverse set of the OOD samples. 
    \item We demonstrate the effectiveness of our approach on the benchmark datasets. The proposed approach out-performs the current state of the art OOD detection methods on almost all the tested cases with a substantial increase of 63\% and 56\% in the true negative rate, i.e. the fraction of the detected OODs, from the softmax scores based method ODIN~\citep{odin} and the Mahalanobis distance from the closest class based method~\citep{mahalanobis} respectively.
\end{itemize}
}}



\noop{
In this paper, we systematically study the diversity in the nature of the OODs and their sources of uncertainty. We examine how different detection methods address different kinds of OODs. Figure~\ref{fig:OOD_types} illustrates the  different types of OODs. We exploit this to develop a new integrated approach to OOD detection that outperforms the existing state of the art methods. 
Our three central contributions are: 
\begin{itemize}[noitemsep, topsep=0pt, leftmargin=*]
    \item Defining a taxonomy of OODs and examining how state of the art detection methods fail to detect different types of OODs. 
    \item Using our observations to develop a novel algorithm \RK{approach?} that can detect all types of the defined OODs. 
    \item Demonstrating its effectiveness on benchmarks such as MNIST, CIFAR10 and SVHN as in-distribution (ID) data with different DNN architectures and several OOD datasets. The proposed approach out-performs the current state of the art OOD detection methods on almost all the tested cases with a substantial increase of 63\% and 56\% in the true negative rate, i.e. the fraction of the detected OODs, from the softmax scores based method ODIN~\citep{odin} and the Mahalanobis distance from the closest class based method~\citep{mahalanobis} respectively.
\end{itemize}
}

\noop{

\section{Introduction}
\label{sec:intro}
Deep neural networks (DNNs) have achieved near human-level accuracy in many domains 
such as computer vision~\citep{img-classification}, speech recognition~\citep{speech-recog}, and text analysis~\citep{text-analysis}. But their deployment in the safety-critical systems 
such as self-driving vehicles~\citep{ML-App7}, aircraft collision avoidance~\citep{NNinAircraft}, and medical diagnoses~\citep{NNclinically} is hindered by their brittleness and the resulting lack of trust.
One major challenge is the inability of DNNs to be self-aware of when these models are outside their training distribution and likely to produce incorrect predictions. It has been widely reported in literature~\cite{guo2017calibration,baseline} that deep neural networks overfit in the negative log likelihood space and consequently exhibit overconfident predictions even on inputs which are from a different distribution far from the training data and likely to produce wrong predictions. The responsible deployment of deep neural network models in high assurance applications necessitates detection of  out-of-distribution (OOD) data so that DNNs can abstain from making decisions on those. 


A number of techniques have been recently proposed for the detection of OODs or calibration of the DNN's confidence. 
These methods use different statistical, geometric or topological  signatures that  
differentiate OODs from the samples in the training distribution. Some examples of these techniques include the use of conformance measure amongst the labels of the nearest neighbors~\citep{dknn}, Mahalanobis distance from the closest class~\citep{mahalanobis}, distribution of the softmax scores~\citep{kl-div, baseline, odin, temp-scaling}, topology-aware manifold learning~\citep{jang2019need}, reconstruction error~\citep{vae-recon-err},  likelihood-ratio~\citep{likelihood-ratio}, density estimates~\citep{towardsNN} and trust scores (ratio of the distance to the nearest class different from the predicted class and the distance to the predicted class)~\citep{trust-score}. These methods attempt to present an uniform approach to detect all OODs accompanied with empirical evaluation. Our study shows that there are diverse classes of outliers which are amenable to different techniques. 

\section{Main Results}
In this paper, we systematically study the diversity in the nature of OODs with respect to their sources of uncertainty and examine how different detection methods address different kinds of OODs. We exploit this study to develop a new integrated approach to OOD detection that outperforms all existing state of the art methods. 
The three central contributions of this paper are: 
\begin{itemize}[noitemsep, topsep=0pt]
    \item identifying a taxonomy of OODs,
    \item examining how  state of the art detection methods fail to detect all types of OODs,
    \item using our observations to propose a novel OOD detection algorithm, and
    \item demonstrating its effectiveness on benchmarks such as MNIST, CIFAR10 and SVHN as in-distribution data with different DNN architectures and several out-of-distribution datasets.
\end{itemize}

\begin{wrapfigure}{r}{0.55\textwidth}
  \begin{center}
    \includegraphics[scale=.4]{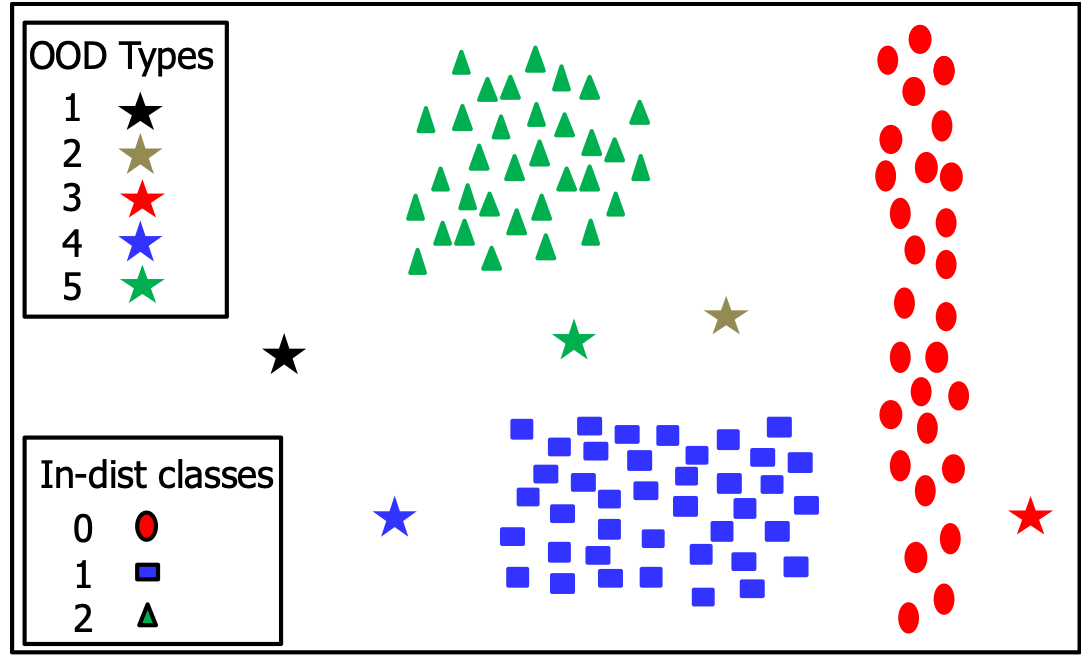}
  \end{center}
  \caption{The different types of OODs ( shows as stars) in a 2D space with three different classes.
  }
\label{fig:OOD_types}
\end{wrapfigure}
\noindent \textbf{OOD Taxonomy}. Figure~\ref{fig:OOD_types} illustrates the  different types of OODs.
Type 1 OOD sample has high epistemic uncertainty and is away from all training data of all classes. Type 2 OOD has high epistemic uncertainty with respect to each of the three classes even though approximating distribution of all training data together using simple kernels such as Gaussian would not find it to be an outlier. Type 3 OOD has high aleatoric uncertainty 
  and lies close to the decision boundary between class 0 and class 1. 
  Type 4 and 5 have high epistemic uncertainty with respect to the class predicted  by the DNN model. While Type 4 OOD is far from the distribution along the principal axis. Type 5  OOD varies along the non-principle axis which has very low variance and even a relatively small deviation in this axis indicates it is OOD. In Section~\ref{sec:taxonomy}, we demonstrate how existing methods fall short of detecting all types of OODs. 

\noindent\textbf{Integrated OOD Detection Approach.} We identify strength weakeness of each approach - to summarize here -- then framework - something Something something Something something Something something Something something Something something Something something Something something Something something Something something Something something Something something Something something Something something Something something Something something Something something Something something Something something Something something Something something Something something Something something Something something Something something Something something Something something Something something Something something Something something Something something Something something Something something Something something Something something Something something Something something Something something Something something Something something Something something Something something Something something Something something Something Something something Something something Something something Something something  

\noindent\textbf{Experimental Evaluation} Something something Something something Something something Something something Something something Something something Something something Something something Something something Something something Something something Something something Something something Something something Something something Something something Something something Something something Something something Something something Something something Something something Something something Something something Something something Something something Something something Something something Something something Something something Something something Something something Something something Something something 

\noop{
OODs of type 1 lie very close to the in-distribution samples of classes 
0 and 1.    
So, they are non-conformant in the labels of their nearest neighbors from the training samples. OODs of type 2 and 3 are conformant in the labels of their nearest neighbors (class 1 and 2 respectively). They, however vary in the magnitude of the reconstruction error from the principal component of the nearest class. This error will be lower(higher) for OODs of type 3(2) as they lie(do not lie) along the principal component of their nearest class. OODs of type 4 and 5 are similar to each other and OODs of type 1 in terms of the non-conformance in the labels of the nearest neighbors from the training samples. They, however vary in their density estimates used for OOD detection. OODs of type 5 will be far from the density estimate of the in-distribution if this estimate is made for each class separately. On the other, if a single over-estimation is used for the entire in-distribution, then OODs of type 5 would lie closer to space of the in-distribution samples. Irrespective of the type of in-distribution density estimate, OODs of type 5 will always be far from this estimate. On the other hand, OODs of type 1 will always have their density estimates very close to the in-distribution classes (0 or 1).

\begin{figure}[!h]
\centering
\includegraphics[scale=.7]{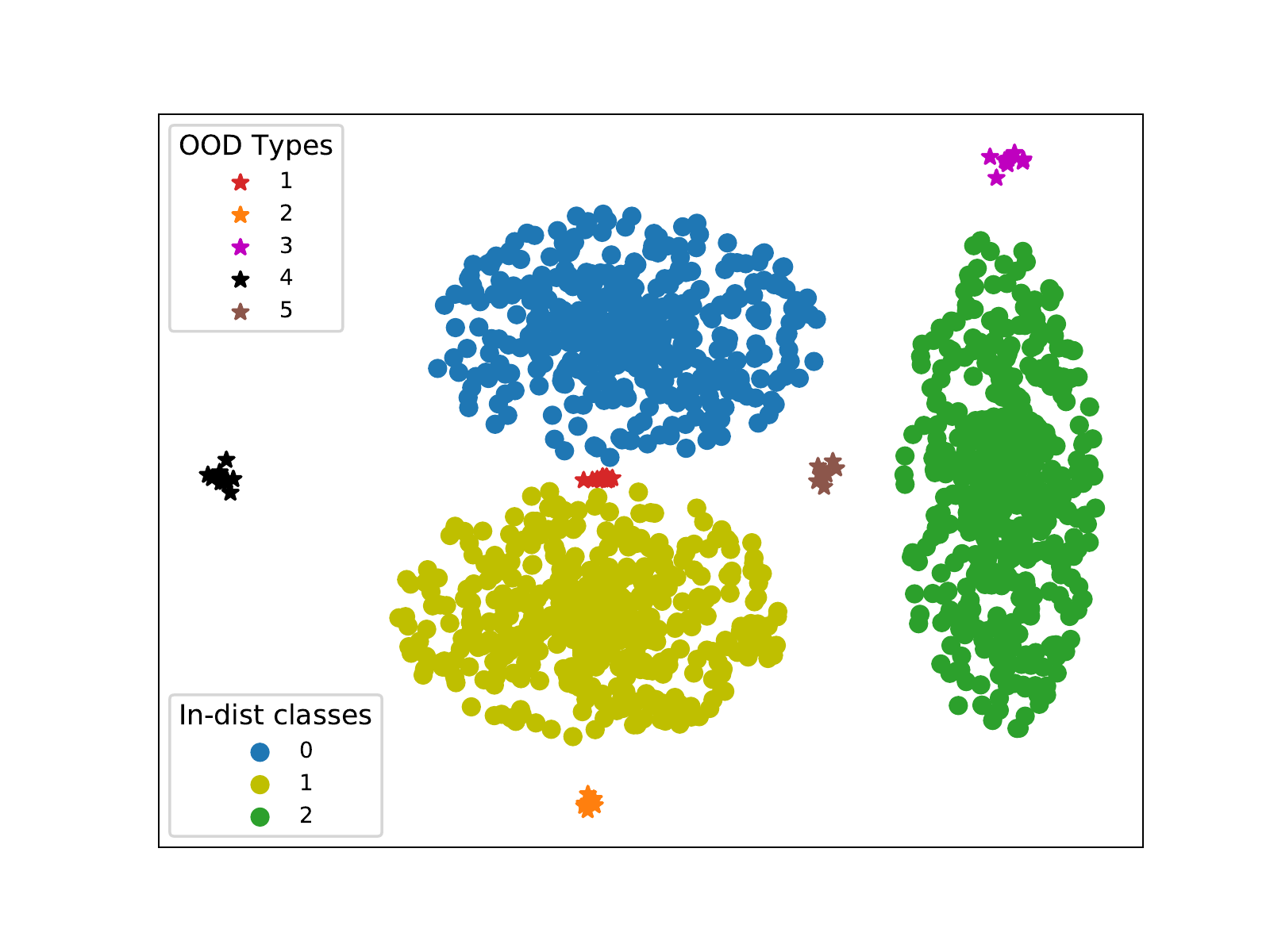}
\caption{Different types of OODs}
\label{fig:OOD_types}
\end{figure}
 Since different types of OODs differ in their distinguishing properties from the in-distribution samples, using only one type of these distinguishing properties might not be able to detect those OODs which are similar to the in-distribution with respect to the used property. 

We classify OODs into three classes based on the type of in-distribution (tied or class-wise) and the sources of uncertainty (aleatory and epistemic) in the prediction of a classifier. We also provide a hypothesis on where and why uniform approaches to detect all OODs would fail in the detection of different classes of OODs and validate this hypothesis via experimentation on a simple example. Complementary information about different classes of OODs can be combined to render an OOD detector that is capable of detecting a wider range of OODs. We propose an OOD detector that uses an ensemble of distinguishing signatures to empower the detector with the ability to detect different classes and hence a wider range of OODs.


The contributions of the paper are as follows:
\begin{itemize}
    \item We present a taxonomy of OOD samples and explain how different approaches for OOD detection differ in their ability to detect different classes of OOD samples in Section~\ref{sec:taxonomy}.
    \item 
    \item 
\end{itemize}
}

The rest of the paper is organized as follows. We present a taxonomy of OOD samples that identifies mutually non-exlcusive types of OODs and describes how different OOD detection approaches differ in their ability to detect different classes of OOD samples in Section~\ref{sec:taxonomy}. We exploit the identified taxonomy to develop a common framework in Section~\ref{sec:framework} based on modeling distribution of internal hidden features of a deep neural network on training data, and measuring deviation from this distribution during inference to detect OOD samples. We experimentally demonstrate the effectiveness of this common framework in Section~\ref{sec:exp} and identify how it improves upon its different components before concluding in Section~\ref{sec:conclusion}. 

Compare to OOD detectors that use ensemble of classifiers.

Compare to the survey paper- https://ieeexplore.ieee.org/stamp/stamp.jsp?arnumber=9144212&tag=1 It has tried to list down the strengths and weakness of OOD and adversarial detectors

There is another paper that uses two types of distinguishing features for detection of adversarial examples-  logistic regression classifier with two features as input: the uncertainty and the density estimate- https://arxiv.org/pdf/1703.00410.pdf 

https://arxiv.org/pdf/1909.11786.pdf - they use Mahalanobis distance measure with class-wise covariance combined with GMM to detect OODs

}

\section{OOD Taxonomy and Existing Detection Methods}
\label{sec:taxonomy}
DNNs predict the class of a new input based on the classification boundaries learned from the samples of the training distribution. 
Aleatory uncertainty is high for inputs which are close to the classification boundaries, and epistemic uncertainty is high when the input is far from the learned distributions of all classes~\citep{uncertainty, ML-uncertainty-survey}. Given the predicted class of a DNN model on a given input, we can observe the distance of the input from the distribution of this particular class and identify it as an OOD if this distance is high. We use this top-down inference approach to detect this type of OODs which are characterized by an  inconsistency in model's prediction and input's distance from the distribution of the predicted class. Further, typical inputs to DNNs are high-dimensional and can be decomposed into principal and non-principal components based on the direction of high variation; this yields another dimension for classification of OODs. We, thus, categorize an OOD using the following three criteria.
\begin{enumerate}[topsep=0pt,leftmargin=*]
    \item Is the OOD associated with higher epistemic or aleatoric uncertainty, i.e., is the input away from  in-distribution data or can it be confused between multiple classes? 
    \item Is the epistemic uncertainty of an OOD sample unconditional or is it conditioned on the class predicted by the DNN model?
    \item Is the OOD an outlier due to unusually high deviation in the principal components of the data or due to small deviation in the non-principal (and hence, statistically invariant) components? 
\end{enumerate}

\begin{figure}
   \begin{center}
    \includegraphics[width=0.55\textwidth]{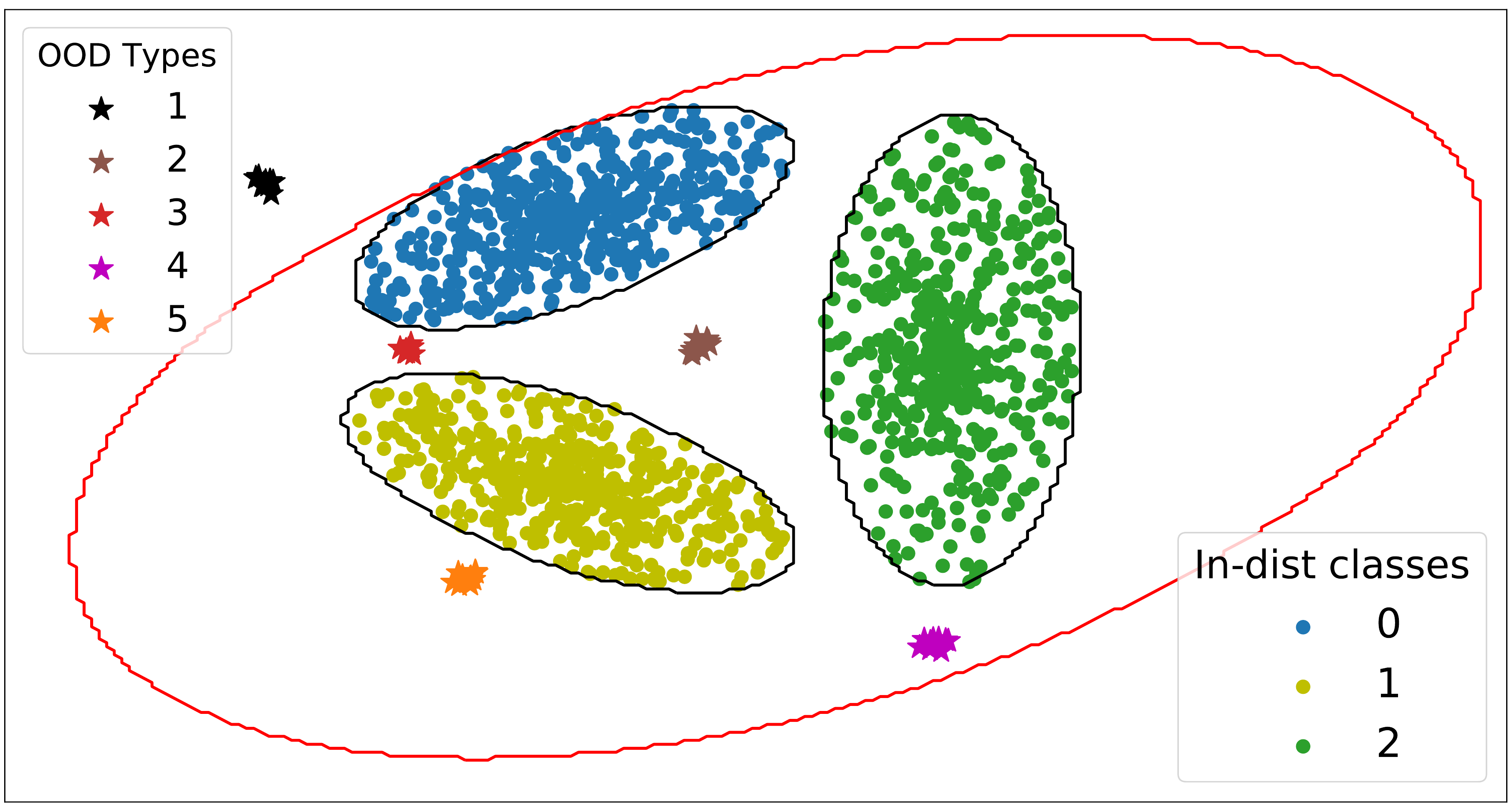}
    \caption{The different types of OODs in a 2D space with three different classes. The class distributions are represented as Gaussians with black boundaries and the tied distribution of all training data is a Gaussian with red boundary.}
    \label{fig:OOD_types}
    \end{center}
\end{figure}
Figure~\ref{fig:OOD_types} demonstrates different types of OODs which differ along these criteria. Type 1 OODs have high epistemic uncertainty and are away from the in-distribution data. Type 2 OODs have high epistemic uncertainty with respect to each of the 3 classes even though approximating all in-distribution (ID) data using a single Guassian distribution will miss these outliers. Type 3 OODs have high aleatoric uncertainty as they are close to the decision boundary between class 0 and class 1. Type 4 and 5 have high epistemic uncertainty with respect to their closest classes. While Type 4 OODs are far from the distribution along the principal axis, Type 5  OODs vary along a relatively invariant axis where even a small deviation indicates that the sample is an OOD.

\begin{figure}[!h]
\begin{subfigure}{0.5\textwidth}
  \centering
  \includegraphics[width=0.8\linewidth]{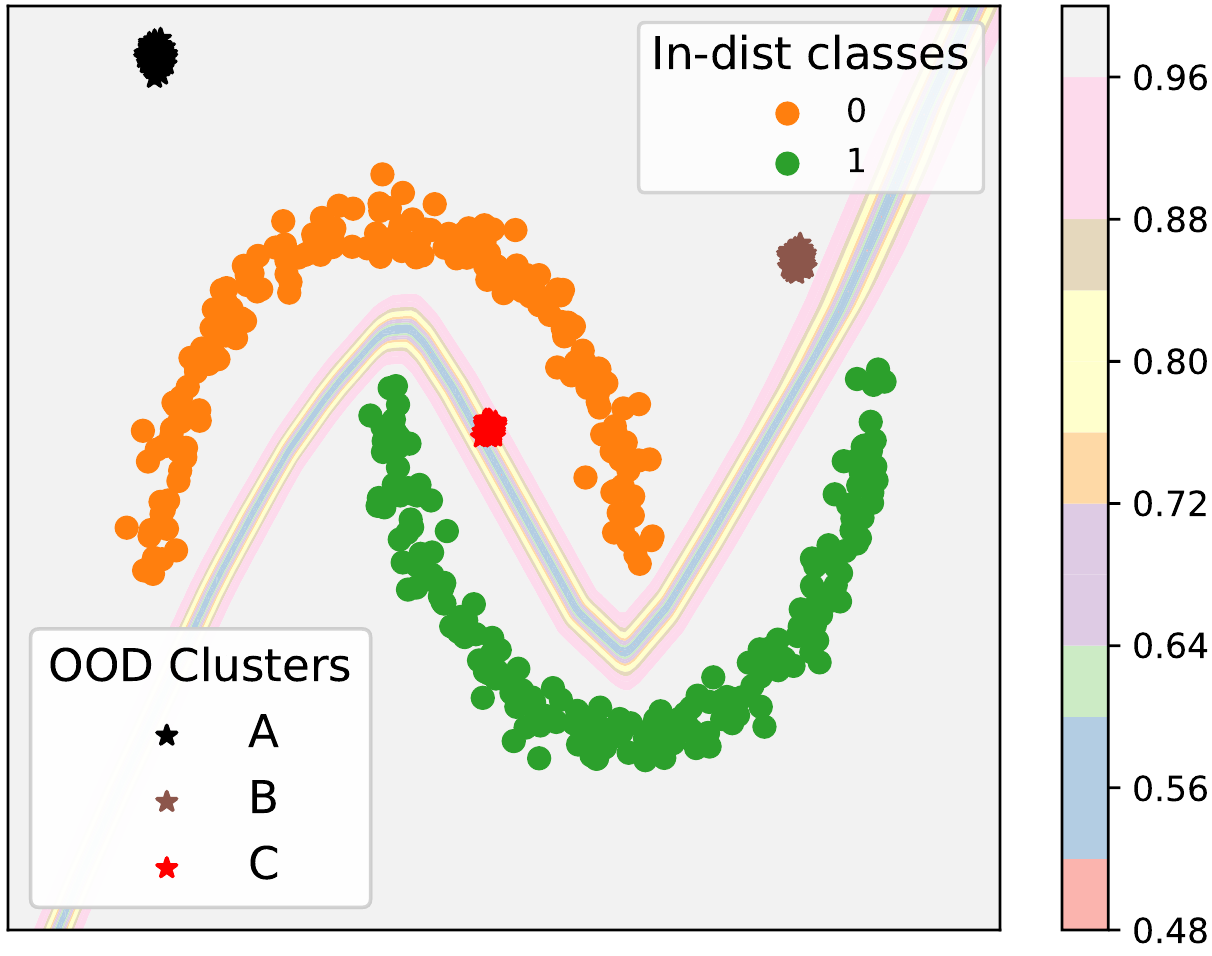}
\end{subfigure}%
\quad
\begin{subfigure}{0.5\textwidth}
  \centering
  \includegraphics[width=0.8\linewidth]{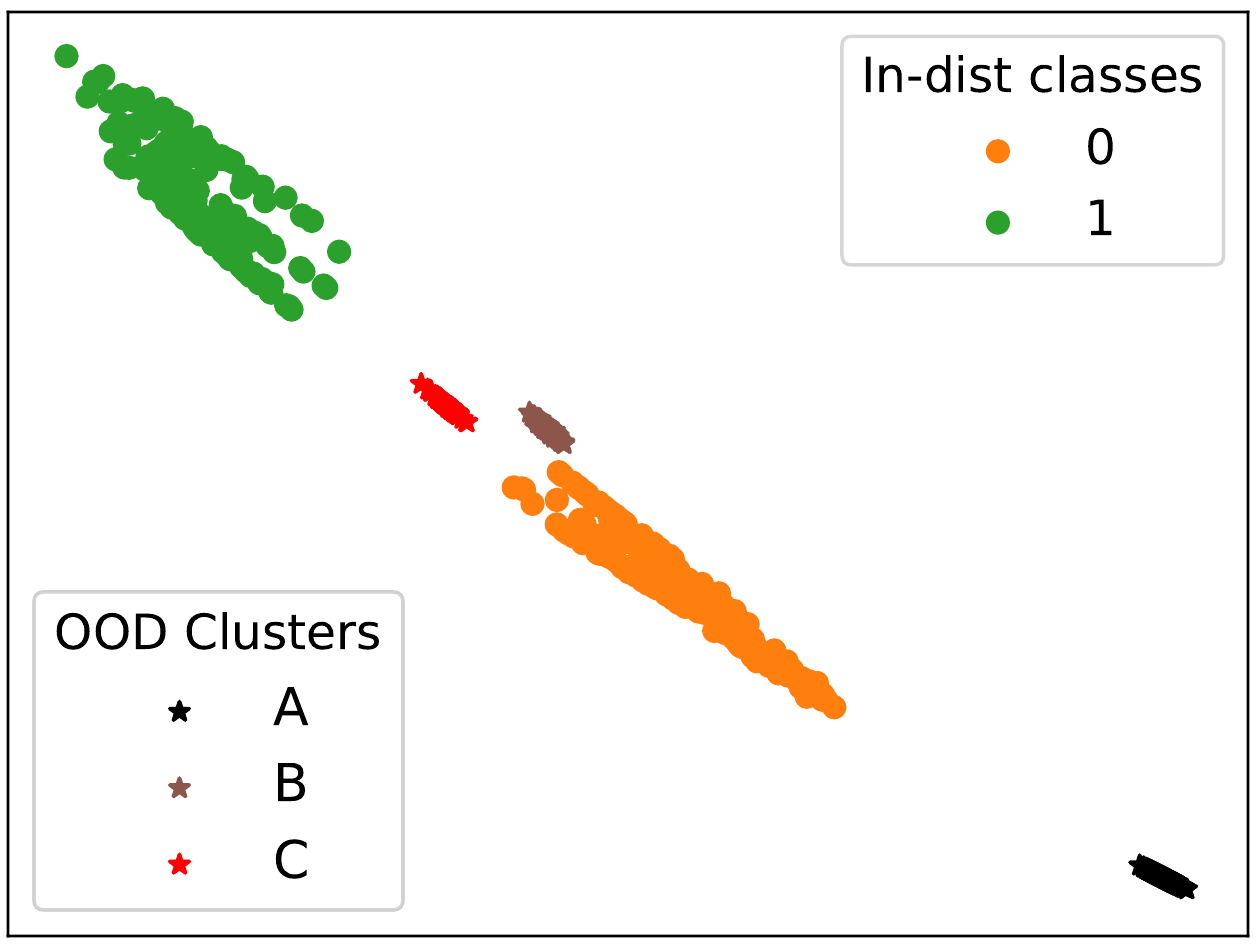}
\end{subfigure}%
\caption{
The cluster A (black), cluster B (brown), and  cluster C (red) clusters represent the OOD due to epistemic uncertainty in the tied training distribution, epistemic uncertainty in the class-conditional training distribution, and the aleatoric uncertainty in the class-conditional distribution, respectively.
(Left) shows the training data of the 2 half-moon classes and the 3 OOD clusters in the input space along with the trained classifier's boundary and its softmax scores. (Right) shows the ID samples and the OODs after projection to the 2D feature space (penultimate layer) of the DNN.} 
\label{fig:toy_ex}
\end{figure}

\indent \textbf{Limitations of Existing Detection Methods.} We empirically demonstrate the limitations of existing OOD detection methods on a two-dimensional (2D) half-moon dataset with two classes. 
As shown in Figure~\ref{fig:toy_ex},
we consider three clusters of OOD samples: cluster A (black), B (brown) and C(red).
Figure~\ref{fig:toy_ex} (right) shows the 2D penultimate features of the classifier. 

\begin{figure}[!h]
    \centering
    \subcaptionbox{Estimating distance from the tied in-distribution fails to detect OOD clusters B and C.}
      {\includegraphics[width=.46\linewidth]{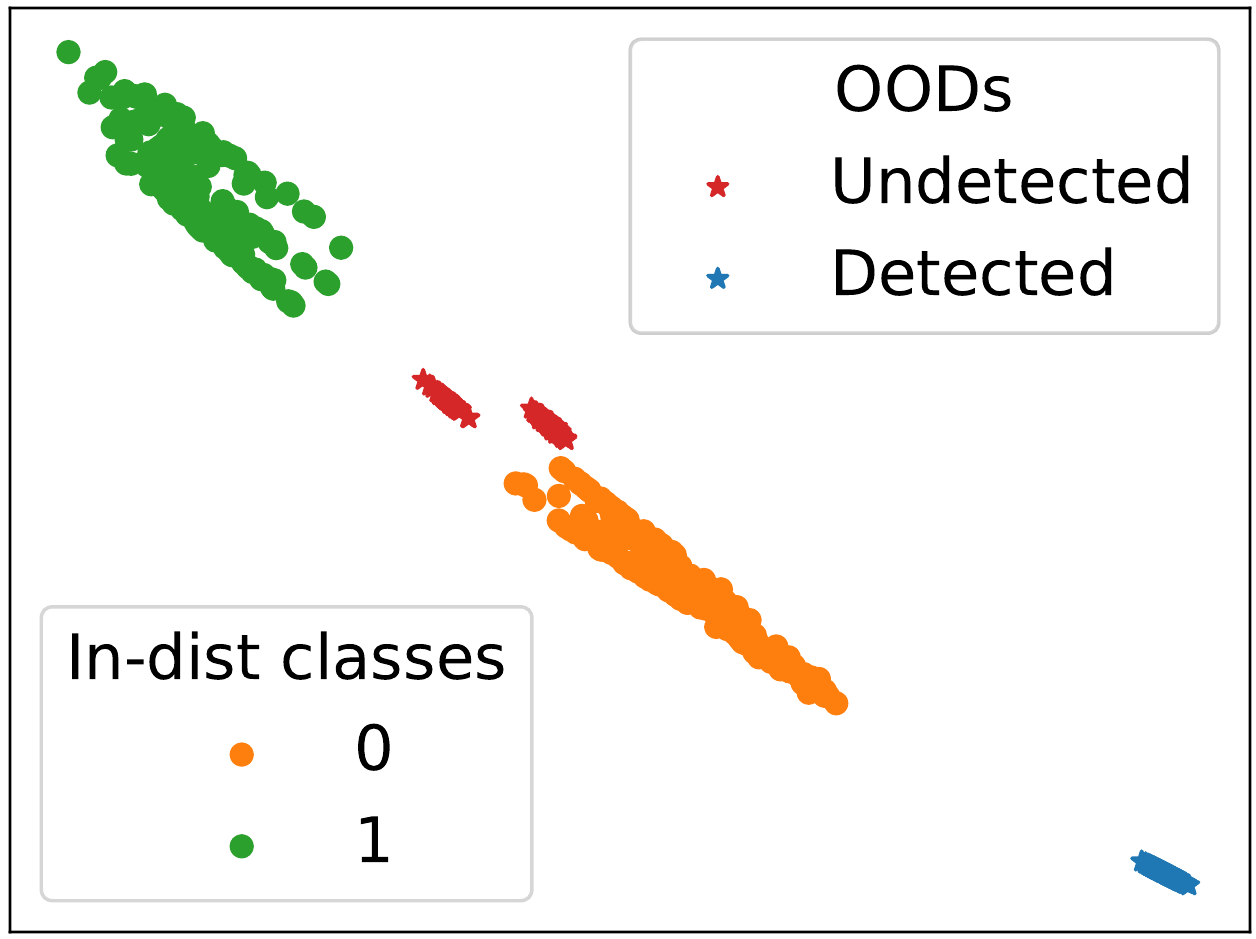}}
  \quad
    \centering
    \subcaptionbox{Techniques based on softmax score fail to detect OODs in cluster A.}
      {\includegraphics[width=.46\linewidth]{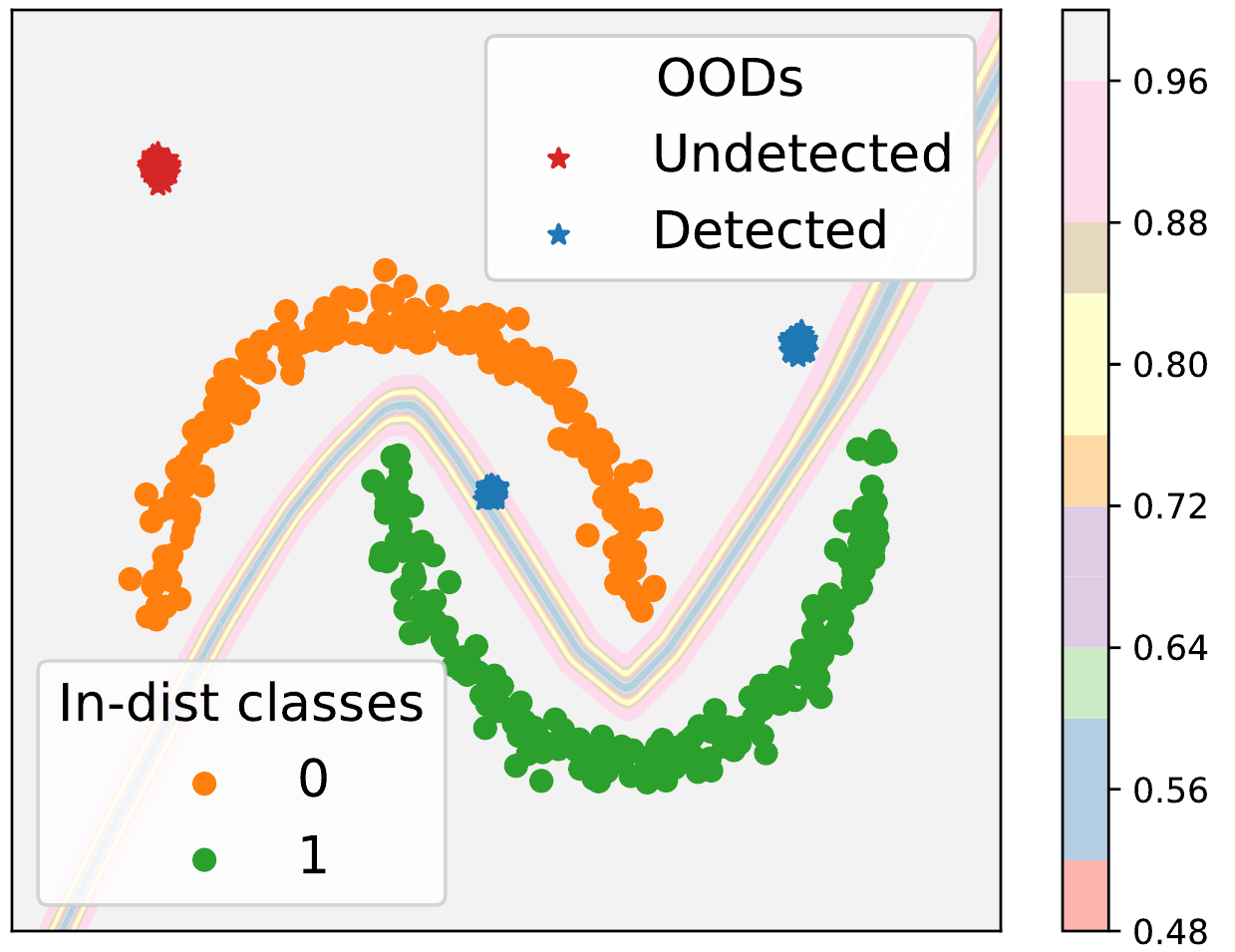}}
    \centering
    \subcaptionbox{Estimating distance from the class-wise in-distribution fails to detect OODs in cluster C.}
      {\includegraphics[width=.46\linewidth]{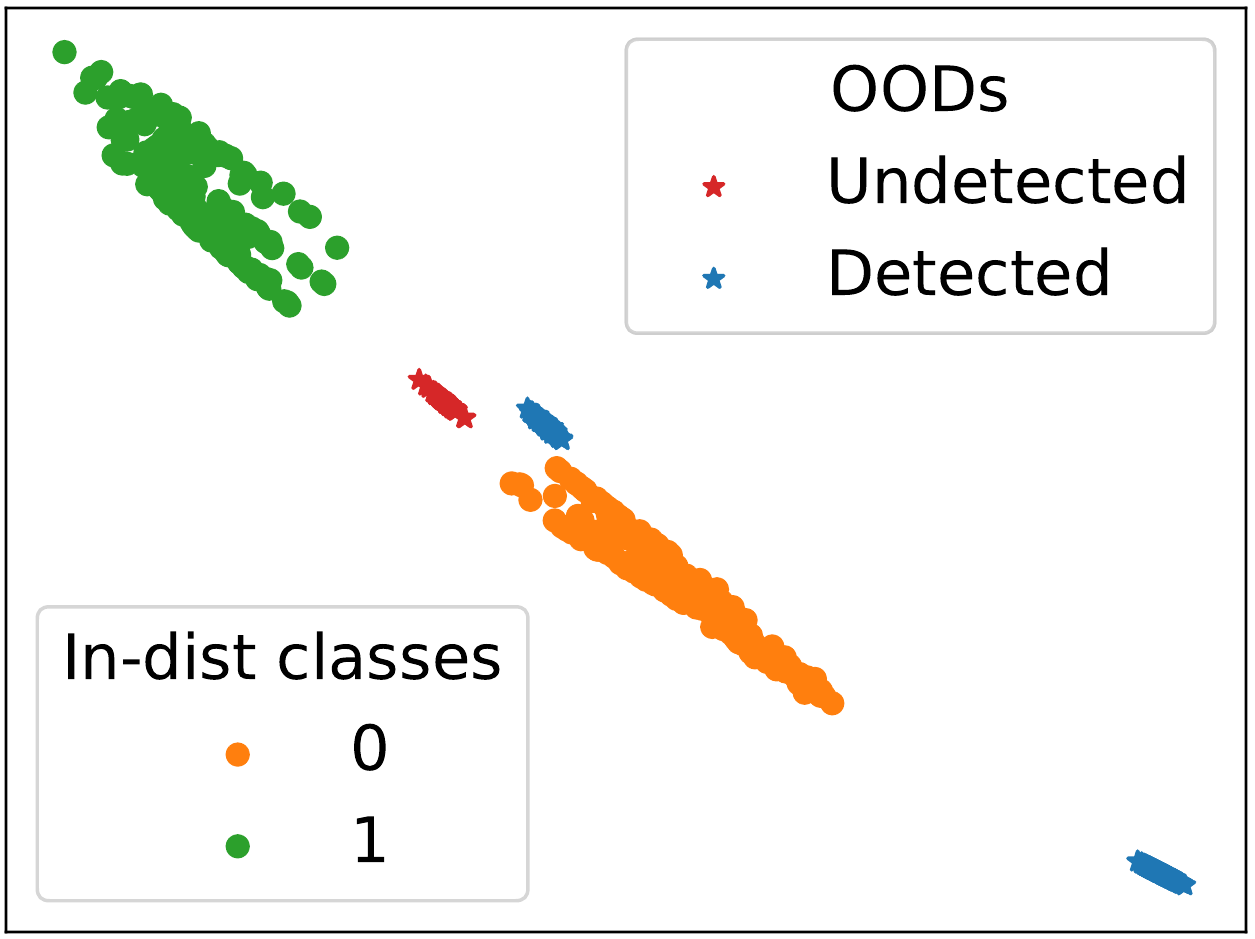}}
  \quad
    \centering
    \subcaptionbox{Non-conformance among the nearest neighbors fails to detect OODs in cluster A and B.}
      {\includegraphics[width=.46\linewidth]{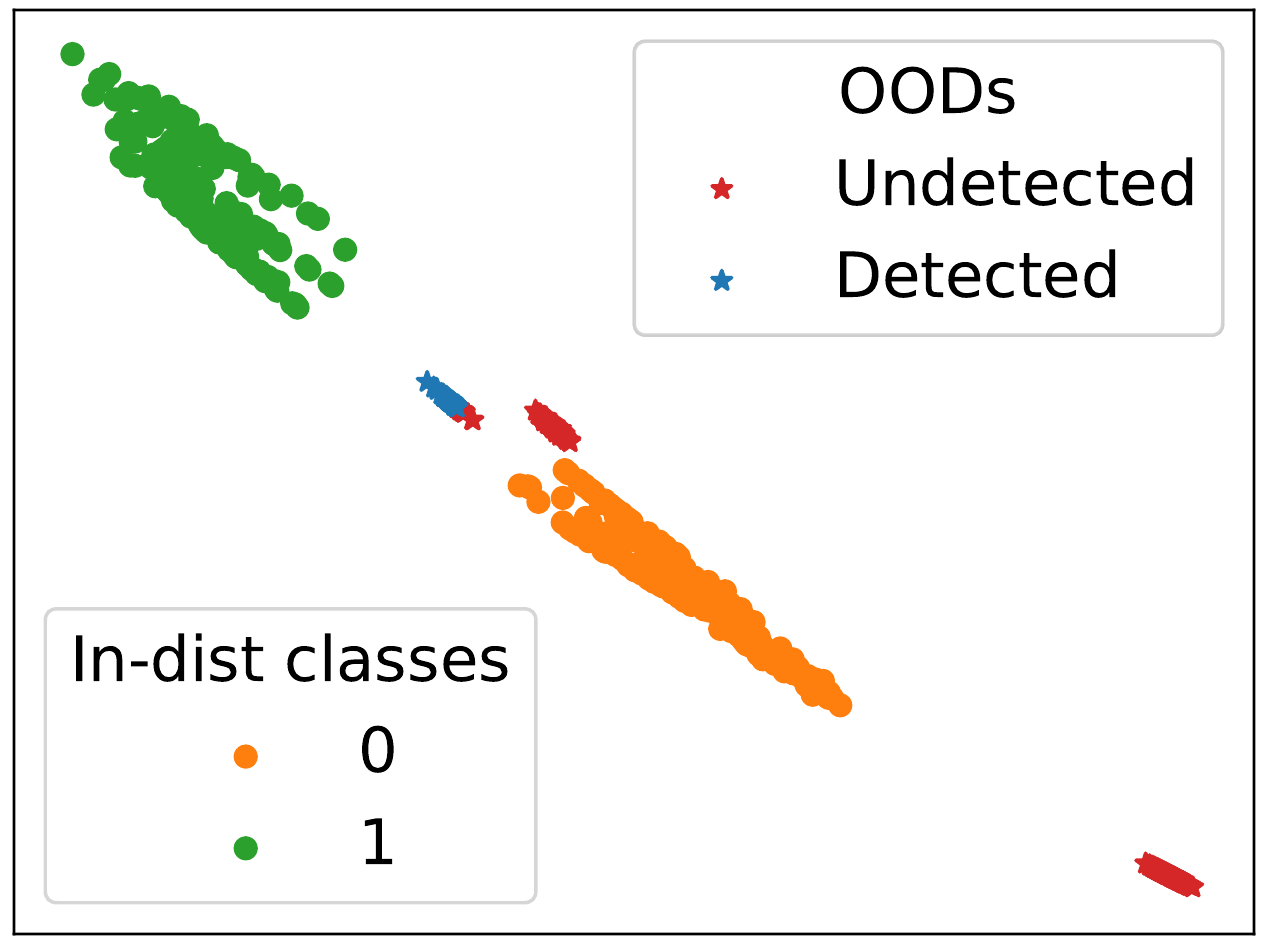}}
\caption{Detected OODs are shown in blue and undetected OODs are in red. Different techniques fail to detect different types of OODs.
}
\label{fig:failure_toy_ex}
\end{figure}

Different approaches differ in their ability to detect different OOD types as illustrated in Figure~\ref{fig:failure_toy_ex}.
\begin{itemize}[topsep=0pt,leftmargin=*]

    \item Figure~\ref{fig:failure_toy_ex}(a) shows that the  Mahalanobis distance~\citep{mahalanobis} from the mean and tied covariance of all the training data in the  feature space cannot detect
    OODs in the clusters B and C corresponding to 
    class-conditional epistemic uncertainty and aleatoric uncertainty, respectively. It attains the overall true negative rate (TNR) of 39.09\% at  the 95\% true positive rate (TPR). 
    
     \item Figure~\ref{fig:failure_toy_ex}(b) shows that the softmax prediction probability (SPB)~\citep{baseline} cannot  detect the OODs in cluster A corresponding to high epsitemic uncertainty. The TNR ( at 95\% TPR) reported by the SPB technique is 60.91\%.  
     \item Figure~\ref{fig:failure_toy_ex}(c) shows that class-wise Principal Component Analysis (PCA)~\citep{pca} cannot detect OODs in cluster C corresponding to high aleatoric uncertainty. We performed PCA of the two classes separately in the  feature space and used the minimum reconstruction error to detect OODs. 
     This obtained overall TNR of 80.91\% (at 95\% TPR). 
     \item Figure~\ref{fig:failure_toy_ex}(d) shows that K-Nearest Neighbor (kNN)~\citep{dknn} non-conformance in the labels of the nearest neighbors cannot detect OODs in clusters A and B with high epistemic  uncertainty. 
     The overall TNR (at 95\% TPR) reported by this technique is 15\%. 
\end{itemize} 

These observations can be explained by the focus of different detection techniques on measuring different forms of uncertainty. This motivates our integrated OOD detection method. 
\section{Integrated OOD Detection Method}
\label{sec:framework}
\begin{figure}[!h]
    \centering
      {\includegraphics[width=.35\linewidth]{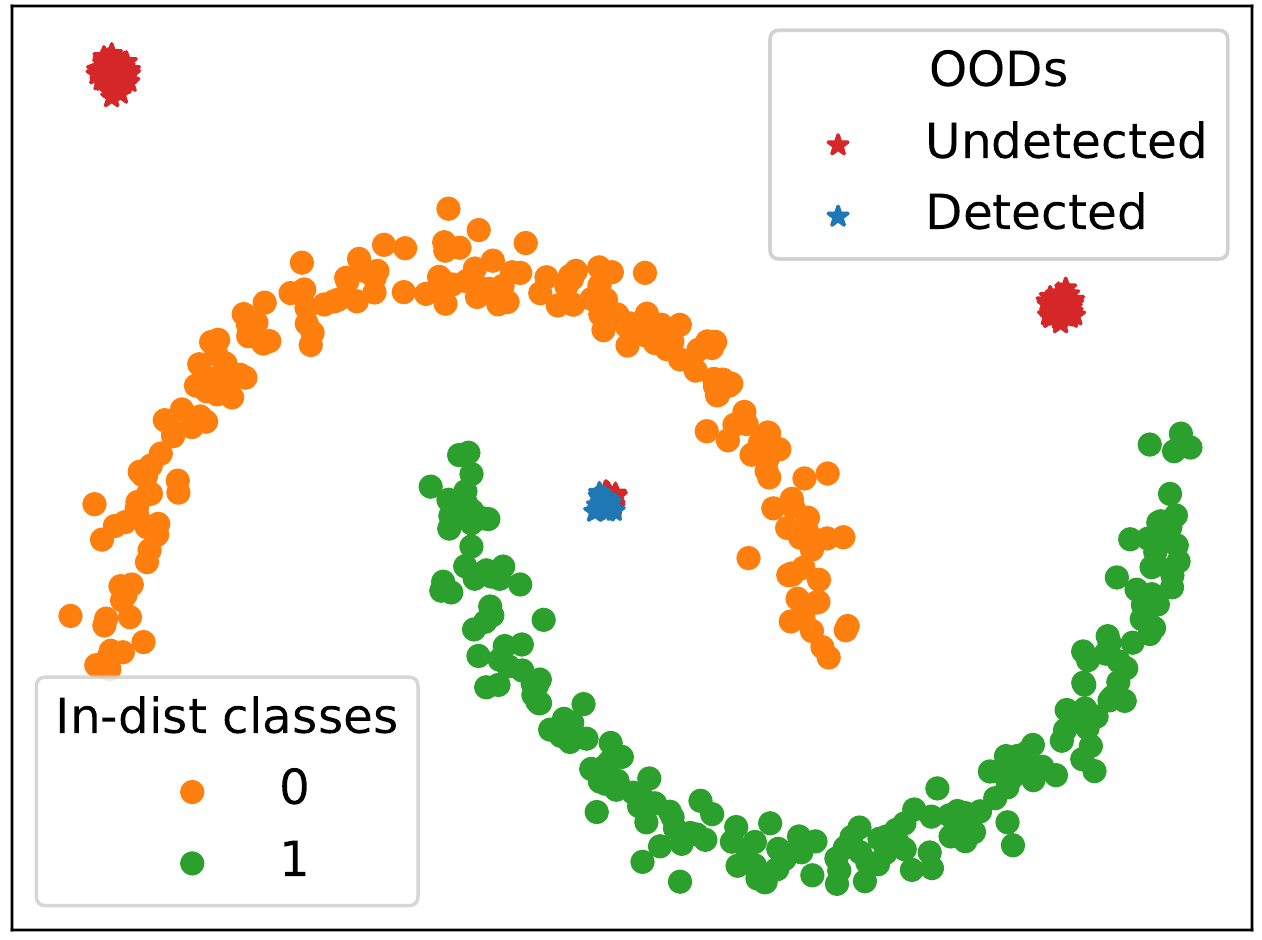}}
    \centering
      {\includegraphics[width=.35\linewidth]{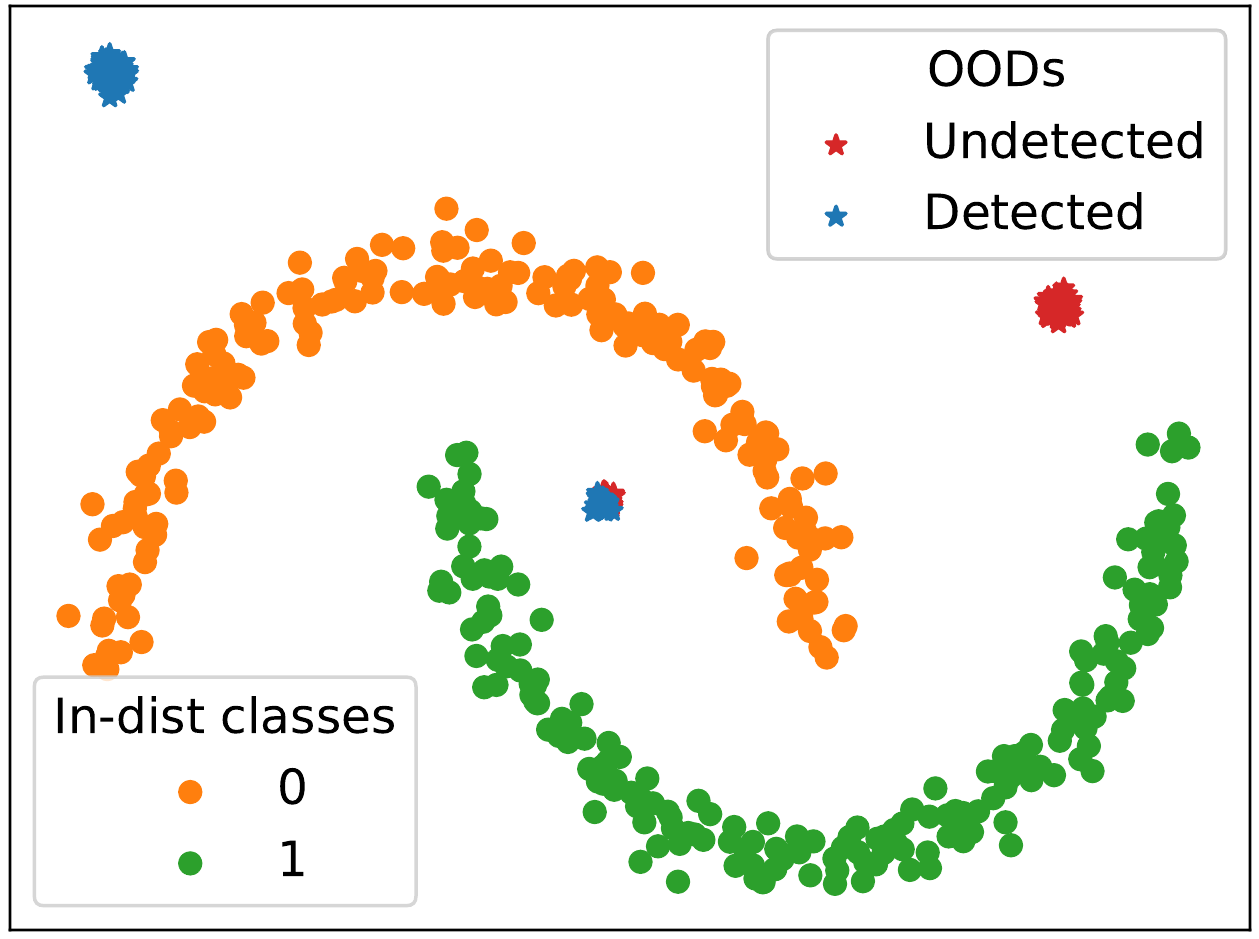}}
    \centering
      {\includegraphics[width=.35\linewidth]{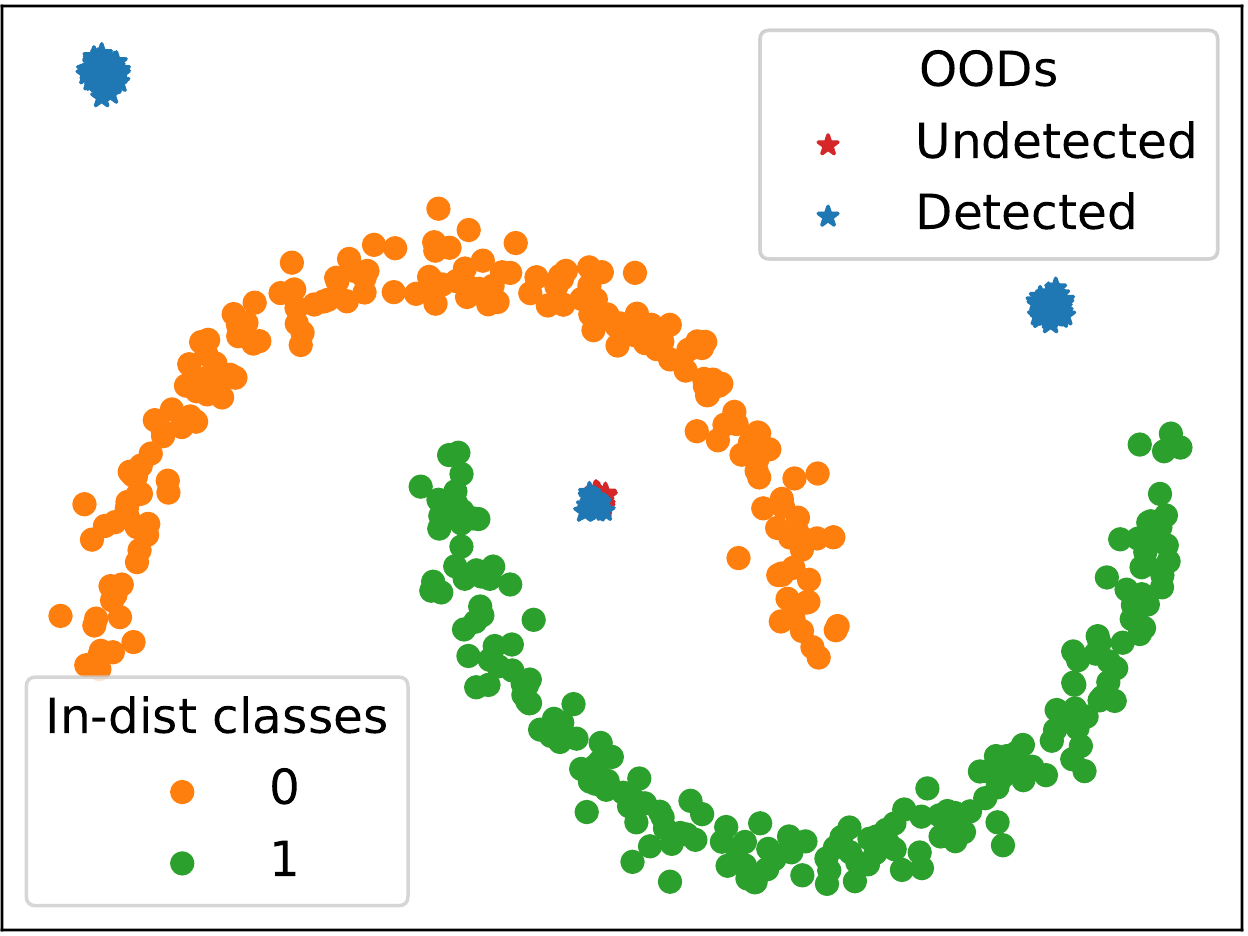}}
    \centering
      {\includegraphics[width=.35\linewidth]{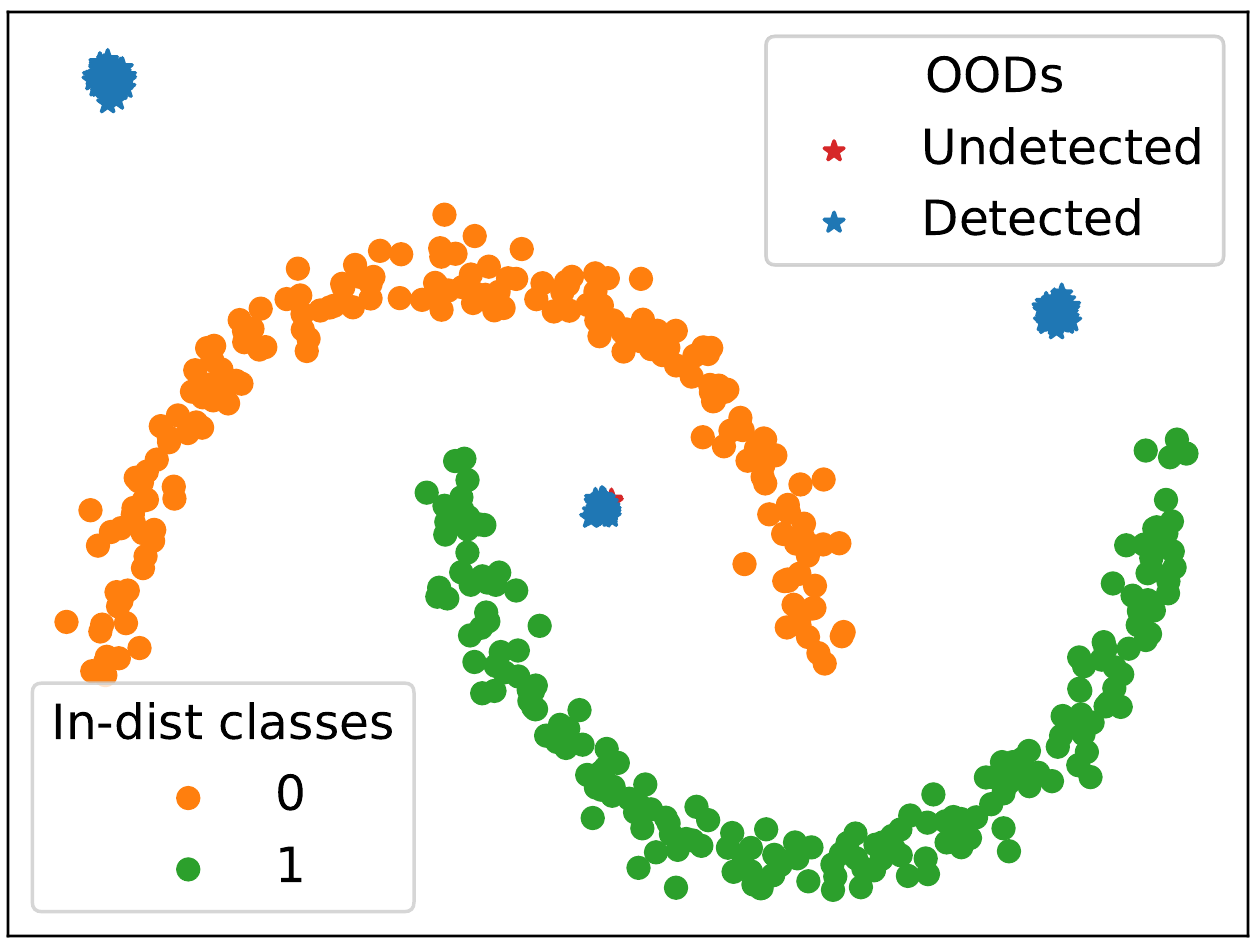}}
\caption{Complementary information about different types of OODs improves detection. (Top-left) \textbf{15\%} TNR with non-conformance among the labels of the nearest neighbors. (Top-right) Adding Mahalanobis distance over the tied in-distribution improves TNR to \textbf{54.09\%}. (Bottom-left) Adding Class-wise PCA further improves TNR to  \textbf{95.91\%} TNR. (Bottom-right) Adding softmax score further improves TNR to \textbf{99.55\%}. TPR is 95\% in all the cases.}
\label{fig:framework_toy_ex}
\end{figure}

Complementary information about different OOD types can be used to detect a wider range of OODs. Figure~\ref{fig:framework_toy_ex} shows the improvement in the TNR of the OOD detector composed with information about different classes of OODs on the two half-moons dataset. Non-conformity in the labels of the nearest neighbors captures OODs in cluster C. Mahalanobis distance from the tied in-distribution detects OODs in cluster A. Reconstruction error from the PCA of the 2 class distributions captures OODs in cluster B. Softmax scores further strengthens the OOD detection by reporting OODs in cluster C that are undetected by the other three methods.

\noop{
If $\newinp$ is an OOD, we can identify it answering the following three questions: 

\begin{enumerate}[noitemsep,topsep=0pt,leftmargin=*]
    \item Is the OOD input having higher epistemic or aleatoric uncertainty, that is, is the input away from all training data or is it unusually more confusing? 
    \\ \RK{The answer for the epistemic is given by the Mahalanobis distance (with class-wise mean and tied or class-wise variance). The answer for the aleatoric is given by the NN conformance and the softmax score. Do we need to answer why are we using 2 features for answering aleatoric uncertainty question? If so, then we can say that as observed from the experiment on toy dataset, softmax adds complementary strength to the NN method by detecting those OODs which NN missed.}
    \item Is the epistemic uncertainty of the OOD input unconditional or it it conditioned on the class predicted by the DNN model?
    \item Is the OOD an outlier due to unusually high deviation in the principal components of the data or due to small deviation in the non-principal (and hence, invariant) components? 
    \\ \RK{The answer for 2 and 3 are given by the reconstruction error of the PCA in the class-conditional distribution of the training data.}
\end{enumerate}
}

The integrated OOD detection approach, thus, uses the following attributes, each specialized in detecting a specific type (or a combination of types) of OODs:

\begin{enumerate}[leftmargin=*]
    \item  Mahalanobis distance from the in-distribution density estimate that considers either tied~\citep{mahalanobis} or class-wise covariance estimate. This attribute captures  the overall or class-conditional epistemic uncertainty of an OOD. Our refinement to also use class-wise covariance significantly improves detection of OODs when coupled with PCA approach described below. 
    \item Conformance measure among the variance of the Annoy~\citep{annoy} nearest neighbors
    calculated as the Mahalanobis distance of the input's conformance to the closest class conformance. Our experiments found this to be very effective in capturing aleatoric uncertainty. This new attribute is a fusion of nearest-neighbor and Mahalanobis distance methods in literature.
    \item Prediction confidence of the classifier as the maximum softmax score on the perturbed input where the perturbation used is the same as ODIN approach~\citep{odin}. This boosts the detection of high aleatoric uncertainty by sharpening the class-wise distributions. 
    \item Reconstruction error using top  40\% of PCA components where the components are obtained via class conditional PCA of the training data. This boosts the detection of high class-wise epistemic uncertainty by eliminating irrelevant features. 
\end{enumerate}

This fusion of attributes from existing state-of-the-art detection methods and new attributes was found to be the most effective integrated appraoch capable of detecting the different types of OODs. We  evaluated it on several benchmarks as discussed in Section~\ref{sec:exp} with ablation study in Appendix.

\noop{
While mahalanobis distance estimation with tied covariance and SPB attributes are used by the existing state-of-the-art OOD detection methods, we introduce two new attributes based on the mahalanobis distance estimate with class covariance and conformance measure among the variance of nearest neighbors to enhance the detection on all identified types of OODs by capturing both epistemic and aleatoric uncertainty.

These attributes are selected such that the framework satisfies the following two properties: 
\begin{enumerate}
    \item \textbf{Sound:} This property is with respect to the choice of attributes in the framework. Each attribute should be able to differentiate in-distribution samples from at least one class of OODs from its taxonomy.
    \begin{equation}
        \forall \text{attribute} \in \text{framework}, \exists \text{OOD type} \in \text{taxonomy},
        \text{s.t. attribute captures differentiating property of the OOD type}...
    \end{equation}
    \item \textbf{Complete:} This property is with respect to the three classes of OODs defined in its taxonomy. Each class of OOD should have at least one attribute in the framework that tries to differentiate in-distribution samples from OODs belonging to that class.
    \begin{equation}
            \forall \text{OOD type} \in \text{taxonomy}, \exists \text{attribute} \in \text{framework}, \text{s.t. the property differentiating the OOD type is captured by the attribute}
    \end{equation}
\end{enumerate}


The weighted sum of these $n$ attributes in the framework forms a signature differentiating in-distribution samples from a diverse set of OODs:

\begin{equation}
    Signature(OOD\ detector) = \sum_{i=1}^n w_i \times \text{attribute}_{i}
\end{equation}

As illustrated in the experimental section, an OOD detector based on this signature out-performs the state of the art OOD detectors by capturing a wider range of OODs on several benchmarks.

} 
\section{Experimental Results}
\label{sec:exp}
\textbf{Attributes forming the signature of the OOD detector used in the experiments}
The signature of the OOD detector used in the experiments is the weighted sum of four attributes, one from each of the following four categories:
\begin{enumerate}
    \item \textbf{Distance from the in-distribution density estimate:} We use mahalanobis distance of the input with respect to the closest class conditional distribution. The parameters of this distance are chosen from one of the following two categories:
    \begin{itemize}
        \item empirical class means and tied empirical covariance of training samples
        \item empirical class means and empirical class covariance of training samples
    \end{itemize}
    \item \textbf{Reconstruction error:} We perform class conditional PCA empirically from the training samples. We use minimum reconstruction error of the input from the top 40\% eigen vectors of the class conditional eigen spaces.
    \item \textbf{Prediction confidence of the classifier:} We use maximum value of the temperature scaled softmax scores ($S$) on the perturbed input. Perturbations to the input ($x$) are made according to the following equation~\citep{odin}
    \begin{equation}
        \widetilde{x} = x - \epsilon\text{sign}(-\nabla_{x} \text{log} S_{\hat{y}(x;T)})
    \end{equation}
    The values of the magnitude of noise ($\epsilon$) and the temperature scaling parameter ($T$) are chosen from one of the following three categories:
    \begin{itemize}
        \item $\epsilon = 0$ and $T=0$
        \item $\epsilon = 0$ and $T=10$
        \item $\epsilon = 0.005$ and $T=10$
    \end{itemize}
    \item \textbf{Conformance measure among the nearest neighbors:} We compute an m-dimensional feature vector to capture the conformance among the input's nearest neighbors in the training samples, where m is the dimension of the input. We call this m-dimensional feature vector as the conformance vector. The conformance vector is calculated by taking the mean deviation along each dimension of the nearest neighbors from the input. We hypothesize that this deviation for the in-distribution samples would vary from the OODs due to aleatory uncertainty. 
    
    The value of the conformance measure is calculated by computing mahalanobis distance of the input's conformance vector to the closest class conformance distribution. Similar to the distance for the in-distribution density estimate, the parameters of this mahalanobis distance are chosen from the following two categories:

    \begin{itemize}
        \item empirical class means and tied empirical covariance on the conformance vectors of the training samples
        \item empirical class means and empirical class covariance on the conformance vectors of the training samples
    \end{itemize}
    
    The value of the number of the nearest neighbors is chosen from the set $\{10, 20, 30, 40, 50\}$ via validation. We used Annoy (Approximate Nearest Neighbors Oh Yeah)~\citep{annoy} to compute the nearest neighbors.
\end{enumerate}

The weights of the four attributes forming the signature of the OOD detector are generated in the following manner. We use a small subset (1000 samples) of both the in-distribution and the generated OOD data to train a binary classifier using the logistic loss. The OOD data used to train the classifier is generated by perturbing the in-distribution data using the Fast Gradient Sign attack (FGSM)~\citep{fgsm}. The trained classifier (or OOD detector) is then evaluated on the real OOD dataset at the True Positive Rate of 95\%. The best result, in terms of the highest TNR on the validation dataset (from the training phase of the OOD detector), from the twelve combinations of the aforementioned sub-categories (one from each of the four attributes) are then reported on the test (or real) OOD datasets. 

\textbf{Datasets and metrics.} 
We evaluate the proposed integrated OOD detection on  benchmarks such as  CIFAR10~\citep{CIFAR10} and SVHN~\citep{svhn}. We consider standard metrics~\citep{baseline,odin,mahalanobis} such as the true negative rate (TNR) at 95\% true positive rate (TPR), the area under the receiver operating characteristic curve (AUROC), area under precision recall curve (AUPR), and the detection accuracy (DTACC) to evaluate our performance. 

\textbf{DNN-based classifier architectures.} To demonstrate that the proposed approach generalizes across various network architectures, we consider a wide range of DNN models such as
, ResNet~\citep{resnet}, WideResNet~\citep{wideresnet},
and DenseNet~\citep{densenet}.

\textbf{Comparison with the state-of-the-art.} We compare our approach with the three state-of-the-art approaches: SPB~\citep{baseline}, ODIN~\citep{odin}, and Mahalanobis~\citep{mahalanobis}. For the ODIN method, the perturbation noise is chosen from the set
\{0, 0.0005, 0.001, 0.0014, 0.002, 0.0024, 0.005, 0.01, 0.05, 0.1, 0.2\}, and the temperature $T$ is chosen from the set \{1, 10, 100, 1000\}. These values are chosen from the validation set of the adversarial samples of the in-distribution data generated by the FGSM attack. For the Mahalanobis method, we consider their best results obtained after feature ensemble and input preprocessing with the hyperparameters of their OOD detector tuned on the in-distribution and adversarial samples generated by the FGSM attack.  The magnitude of the noise used in pre-processing of the inputs is chosen from the set \{0.0, 0.01, 0.005, 0.002, 0.0014, 0.001, 0.0005\}.



\indent \textbf{CIFAR10.} With CIFAR10 as in-distribution, we consider SVHN~\citep{svhn}, Tiny-Imagenet~\citep{imagenet}, and LSUN~\citep{lsun} as the OOD datasets. For CIFAR10, we consider two DNNs: ResNet50, and WideResNet. Table~\ref{table:results} shows the results.


\indent \textbf{SVHN.} With SVHN as in-distribution, we consider CIFAR10, Imagenet, and LSUN and as the OOD datasets. For SVHN, we use the DenseNet classifier. 
Table~\ref{table:results} shows the results.


\begin{table}
\begin{center}
\caption{Comparison of TNR, AUROC, DTACC, AUPR with SPB, ODIN and Mahalanobis methods}
\begin{tabular}{ccccccc}
\hline \\


\multicolumn{1}{c}{\bf In-dist} & \multicolumn{1}{c}{\bf OOD Dataset}  &\multicolumn{1}{c}{\bf Method} &\multicolumn{1}{c}{\bf TNR} &\multicolumn{1}{c}{\bf AUROC} & \multicolumn{1}{c}{\bf DTACC} & \multicolumn{1}{c}{\bf AUPR}
\\ \multicolumn{1}{c}{\bf (model)} 


\\ \hline

          \\ CIFAR10\\
          (ResNet50)\\

\\ \hline \\

      & SVHN      & SPB &  44.69	 &\textbf{97.31} 	& 86.36 & 87.78  \\
      &           & ODIN & 63.57 & 93.53 & 86.36 & 87.58 \\
      &           & Mahalanobis & 72.89 & 91.53 & 85.39 & 73.80   \\
      &           & Ours & \textbf{85.90}  &95.14  &\textbf{90.66}  &\textbf{80.01}   \\
      & Imagenet  & SPB &  42.06 	 &90.8	& 84.36 & 92.6  \\
      &           & ODIN & 79.48 & 96.25 &  90.07 & \textbf{96.4}5    \\
      &           & Mahalanobis  &94.26  & \textbf{97.41} & 95.16 & 93.11   \\
      &           & Ours   &      \textbf{95.19} & 97.00 & \textbf{96.02} &90.92    \\
               
      & LSUN  & SPB &  48.37 	 &92.78	& 86.97 & 94.45   \\
      &           & ODIN  & 87.29  & 97.77  & 92.65 &  97.96  \\
      &           & Mahalanobis  & 98.17 & 99.38 & 97.38 & 98.69   \\
      &           & Ours &\textbf{99.36} & \textbf{99.65} & \textbf{98.57} & \textbf{98.96}  \\
\\ \hline

       \\CIFAR10\\
       (WideResNet)\\  
\\ \hline \\

      & SVHN      & SPB  & 45.46 &  90.10 & 82.91 &  82.52  \\ 
      &           & ODIN & 57.14  & 89.30  & 81.14 &  75.48  \\
      &           & Mahalanobis & 85.86 & 97.21 &  91.87 &  \textbf{94.69}  \\
      &           & Ours & \textbf{88.95}  & \textbf{97.61}  & \textbf{92.46} & 92.84  \\
      & LSUN      & SPB & 52.64 & 92.89 & 86.81  & 94.13\\
      &           & ODIN & 79.60 & 96.08 &  89.74  & 96.23 \\
      &           & Mahalanobis & 95.69 & 98.93  & 95.41 & 98.99   \\
      &           & Ours & \textbf{98.84}  & \textbf{99.63} & \textbf{97.72} & \textbf{99.25}\\

\\ \hline\\
         SVHN\\
         (DenseNet)\\
\\ \hline \\
      & Imagenet  & SPB &  79.79 	 &94.78	& 90.21 &97.2 \\
      &           & ODIN & 79.8 &  94.8 & 90.2 & 97.2  \\
      &           & Mahalanobis & \textbf{99.85} &  \textbf{99.88} & \textbf{98.87} & \textbf{99.95}\\
      &           & Ours &   98.02      	 & 98.34	&  98.00 &97.05 \\
      & LSUN  & SPB &  77.12 	 &94.13	& 89.14 &96.96   \\
      &           & ODIN & 77.1 & 94.1 & 89.1 & 97.0\\
      &           & Mahalanobis  & \textbf{99.99} & \textbf{99.91} & \textbf{99.23} & \textbf{99.97}\\
      &           & Ours    	 & 99.74  & 99.79  & 99.08 &99.65   \\
            & CIFAR10   & SPB &  69.31 	 &91.9 	& 86.61  & 95.7 \\
      &           & ODIN & 69.3 & 91.9 & 86.6 & 95.7  \\
      &           & Mahalanobis &  \textbf{97.03} & \textbf{98.92} & \textbf{96.11} & \textbf{99.61}\\
      &           & Ours & 94.87  & 98.41  & 94.97& 98.76  \\
      
\\ \hline \\      
\end{tabular}
\label{table:results}
\end{center}
\end{table}

\indent \textbf{Key observations.} We do not consider pre-processing of the inputs in our integrated OOD detector. Even without input pre-processing and with the exception of CIFAR10 OOD dataset for SVHN in-distribution trained on DenseNet, we could perform equally well (and even out-perform in most of the cases) as the Mahalanobis method on its best results generated after pre-processing the input.

\begin{figure}[t]
  \centering
  \includegraphics[width=1\linewidth]{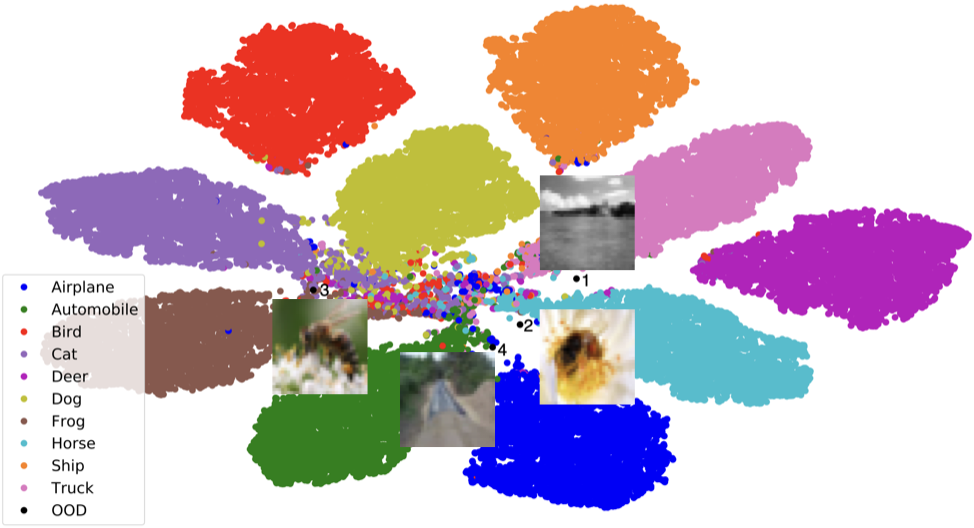}
\caption{t-SNE plot of the penultimate layer feature space of ResNet50 trained on CIFAR10. We show four OOD images from the SCIFAR100.
OOD 1 and OOD 2 are far from the distributions of all classes and thus represent OODs due to epistemic uncertainty. OOD 3 and OOD 4 are OODs due to aleatoric uncertainty as they lie closer to two class distributions. Third OOD is closer to the cat and frog classes of the ID and forth OOD is closer to the airplane and automobile classes of the ID. Mahalanobis distance cannot detect these OODs but our integrated approach can detect them.} 
\label{fig:mahala_our_tsne}
\end{figure}

We also consider a Subset-CIFAR100 as OODs for CIFAR10. Specifically, from the CIFAR100 classes, we select sea, road, bee, and butterfly as OODs which are visually similar to the ship, automobile, and bird classes in the CIFAR10, respectively. 
Thus, there can be numerous OOD samples due to aleatoric and class-conditional epistemic uncertainty which makes the OOD detection challenging. Figure~\ref{fig:mahala_our_tsne} shows the t-SNE~\citep{t-SNE} plot of the penultimate features from the ResNet50 model trained on CIFAR10.  We show 4 examples of OODs (2 due to epistemic and 2 due to aleatoric uncertainty) from Subset-CIFAR100. These OODs were detected by our integrated approach but missed by the Mahalanobis approach. 

These observations justify the effectiveness of integrating multiple attributes to detect OOD samples. 




\indent \textbf{Additional experimental results in the appendix.}  
We also compare the performance of the integrated OOD detector with the SPB, ODIN and Mahalanobis detector in supervised settings, as reported by the Mahalanobis method for OOD detection~\citep{mahalanobis}. These results include experiments  on  CIFAR10,  SVHN and  MNIST  as  in-distribution  data  and  Imagenet,LSUN, SVHN (for CIFAR10), CIFAR10 (for SVHN), KMNIST, and F-MNISTas OOD data across different DNN architectures such as ResNet34, WideResNet,DenseNet, and LeNet5. All these results, along with the ablation studies on OOD detectors with single attributes are included in the Appendix. In almost all of the reported results in the Appendix, our OOD detector could outperform the compared state-of-the-art methods with improvements of even 2X higher TNR at 95\% TPR in some cases.

\noop{
We demonstrate the effectiveness of the proposed integrated OOD detection method on various datasets such as MNIST~\citep{lenet}, CIFAR10~\citep{CIFAR10}, SVHN~\citep{svhn} and different architectures of the DNN based classifiers for these datasets such as Lenet~\citep{lenet}, ResNet~\citep{resnet} and DenseNet~\citep{densenet}. We measure the following metrics: the true negative rate (TNR) at 95\% true positive rate (TPR), the area under the receiver operating characteristic curve (AUROC), and the detection accuracy (DTACC). 

We compare our results with the baseline softmax prediction probability (SPB)~\citep{baseline}, the ODIN~\citep{odin} and the Mahalanobis method~\citep{mahalanobis} to detect OODs. SPB method exploits the difference in the values of the maximum softmax scores for the ID samples(greater) and OODs(lower) to detect OODs. ODIN tries to further separate out the softmax distributions for the ID and OOD samples by adding small perturbations to the input and applying temperature scaling to the softmax scores. Mahalanobis method exploits the difference in the values of mahalanobis distance of the ID(lower) and OOD(higher) samples from the class-wise mean and tied covariance of the ID samples to detect OODs. 

Due to the space limitation, comparison with the SBP method is listed in the appendix. We also performed experiments to compare our results with the other aforementioned methods on the area under the precision-recall curve for both ID (AUIN) and OOD datasets (AUOUT). Results with this metric are also included in the appendix.

\RK{General highlight applicable to all the in-distribution datasets - Performance gain in the TNR over all the other metrics in all the tested cases due to detection of diverse (instead of uniform) types of OODs.}

\indent \textbf{Experiments with the MNIST dataset.} We trained MNIST on LeNet5. With MNIST as the ID dataset, KMNIST~\citep{kmnist} and Fashion-MNIST(F-MNIST)~\citep{fashion-mnist} were used as the OOD datasets. Table~\ref{table:penul-layer-results} shows that our approach
attains a considerably higher TNR on both the OOD datasets in comparison to the ODIN and the Mahalanobis methods.

\indent \textbf{Experiments with the CIFAR10 dataset.} We trained CIFAR10 on three DNN based classifiers: DenseNet, ResNet34, and ResNet50. With CIFAR10 as the ID dataset, STL10~\citep{stl10}, SVHN~\citep{svhn}, Imagenet~\citep{imagenet}, LSUN~\citep{lsun} and a subset of CIFAR100(SCIFAR100)~\citep{CIFAR10} were used as the OOD datasets. While SVHN, Imagenet and LSUN are different from the CIFAR10 dataset, STL10 and the subset of CIFAR100 OOD datasets used in the experiments are quite close to the ID. We used sea, road, bee and butterfly classes from the CIFAR100 dataset as OODs similar to the ship, automobile and birds classes in the CIFAR10 dataset respectively. 

As shown in Table~\ref{table:penul-layer-results}, the integrated approach performed significantly well on the OODs from SVHN, Imagenet and LSUN datasets on the three classifiers. For instance, it could detect 56\% more OODs as compared to the Mahalanobis method from the SVHN OOD dataset with the Resnet50 classifier and 36\% more OODs as compared to the ODIN method on OODs from the LSUN dataset with Resnet34 classifier. With STL10 and SCIFAR100 datasets, the OODs are expected to be close to the class distributions of the CIFAR10 (ID) dataset resulting in OODs due to epistemic and aleatoric uncertainty in the class-conditional distribution of ID. Although the performance of integrated approach was not significant on STL10 dataset (beacuse of the high TPR of 95\%) but it could detect relatively higher percentage of OODs than the other two methods. With SCIFAR100 dataset, our approach could detect considerably(relatively) higher percentage of OODs than the Mahalanobis(ODIN) method. For instance, it could detect 27\% more OODs than the Mahalanobis method on ResNet50 architecture.

\indent \textbf{Experiments with the SVHN dataset.} We trained SVHN on two DNN based classifiers: DenseNet, and ResNet34. With SVHN as the ID dataset, STL10, CIFAR10, Imagenet, LSUN and, SCIFAR100 were used as the OOD datasets. As shown in Table~\ref{table:penul-layer-results}, our OOD detector could out-perform both the ODIN and the Mahalanobis methods on all the OOD datasets with both the architectures. With 99.61\% TNR on SCIFAR100 and ResNet34 architecture, it could detect 63\% and 13\% more OODs as compared to the ODIN and the Mahalanobis method.
}

\noop{
\RK{Highlight- To perform fair comparison, we performed OOD on feature ensemble(features from multiple layers) with our and Mahalanobis approach. Observations-\\
1) Our approach could boost up OOD detection in all cases, and where there was scope to do so...\\
2) Information from multiple layers is also not sufficient to detect those OODs that a method is unable to capture diverse set of OODs. The lower performance of the Mahalanobis method with the feature ensemble on CIFAR10 with Subset-CIFAR100 (Table~\ref{table:all-layers-results}) as the OOD dataset than the integrated method with just the penultimate layer (Table~\ref{table:penul-layer-results}) gives the evidence for this hypothesis.}
}

\noop{
\indent \textbf{Experiments with feature ensemble.} The results listed in Table~\ref{table:penul-layer-results} are with the Mahalanobis and our method on features from the penultimate layer of the classifier. We also compared performance of our method with the Mahalanobis method on features from all the blocks(layers) of ResNet and DenseNet(LeNet) on CIFAR10 and SVHN(MNIST). This approach is referred as feature ensemble in the Mahalanobis paper~\citep{mahalanobis}. The results of these experiments are listed in Table~\ref{table:all-layers-results}. Here also, our method out-performed the Mahalanobis method on almost all the tested cases. 

An important observation made from these experiments is that information from multiple layers is also not sufficient to detect those OODs that a method is unable to detect. The lower performance of the Mahalanobis method with the feature ensemble on CIFAR10 with SCIFAR100 (Table~\ref{table:all-layers-results}) as the OOD dataset than the integrated method with just the penultimate layer (Table~\ref{table:penul-layer-results}) gives the evidence for this hypothesis. 
}

\section{Discussion and future work}
\label{sec:dis}
Recent techniques propose refinement in the training process of the classifiers for OOD detection. Some of these techniques include fine-tuning the classifier's training with an auxiliary cost function for OOD detection~\citep{oe,energy}. Other techniques make use of self-supervised models for OOD detection~\citep{contrastive,self-supervised}. We perform preliminary experiments to compare the performance of these techniques with our integrated OOD detector that makes use of the feature space of the pre-trained classifiers to distinguish in-distribution samples from OODs. Our approach does not require modification of the training cost function of the original task. These results are reported in the Appendix. We consider making use of the feature space of in our OOD detection technique as a promising prospective future work. Another direction of the future work is to explore the score functions used in these refined training processes for OOD detection~\citep{energy,oe,contrastive,self-supervised} as attributes (or categories of attributes) forming the signature of the integrated OOD detector. Another avenue of future work is to explore OOD generation techniques other than adversarial examples generated by the FGSM attack for training of the integrated OOD detector.

\section{Conclusion}
\label{sec:conclusion}
We introduced a taxonomy of OODs and proposed an integrated approach to detect different types of OODs. Our taxonomy classifies OOD on the nature of their uncertainty and we demonstrated that no single state-of-the-art approach detects all these OOD types. Motivated by this observation, we formulated an integrated approach that fuses multiple attributes to target different types of OODs. We have performed extensive experiments on a synthetic dataset and several benchmark datasets (e.g., MNIST, CIFAR10, SVHN).
Our experiments show that our approach can accurately detect various types of OODs coming from a wide range of OOD datasets such as KMNIST, Fashion-MNIST, SVHN, LSUN, and Imagenet. We have shown that our approach generalizes over multiple DNN architectures and performs robustly when the OOD samples are similar to in-distribution data.

\subsubsection*{Acknowledgments}
This work was supported by the Air Force Research Laboratory and the Defense Advanced Research Projects Agency under 
DARPA Assured Autonomy under 
Contract No. FA8750-19-C-0089 and Contract No. FA8750-18-C-0090, 
U.S. Army Research Laboratory Cooperative Research Agreement  W911NF-17-2-0196, 
U.S. National Science
Foundation(NSF) grants \#1740079 and \#1750009,
and the Army Research Office under Grant Number W911NF-20-1-0080. Any opinions, findings and conclusions or recommendations expressed in this material are those of the authors and do not necessarily reflect the views of the Air Force Research Laboratory (AFRL), the Army Research Office (ARO), the Defense Advanced Research Projects Agency (DARPA), or the Department of Defense, or the United States Government.
\clearpage
\bibliography{biblio}

\bibliographystyle{iclr2021_conference}

\newpage
\appendix
\section{Appendix}

\subsection{Defining OODs due to epistemic and aleatoric uncertainty}
\begin{figure}[!h]
\centering
\includegraphics[scale=.45]{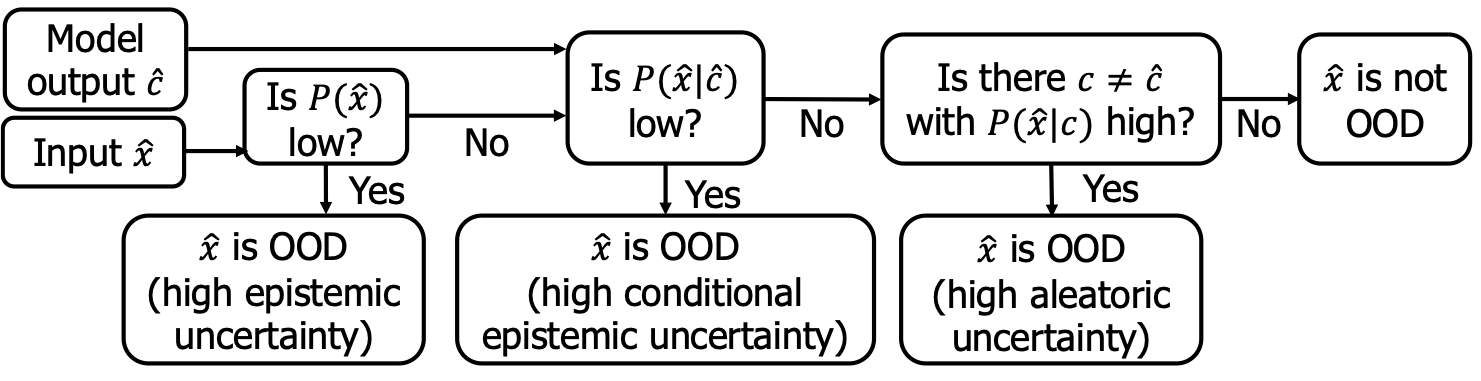}
\caption{OODs due to High Epistemic and Aleatoric Uncertainty} 
\label{fig:OOD_flow}
\end{figure}

In general, let there be $k$ classes $c_1, c_2, \ldots, c_k$ and the distribution of training data for each class is $p(x | c_i)$. The overall training distribution is denoted by $p(x)$. Now, given a new input $\newinp$ to the trained DNN model $M$, let $\newclass = M(\newinp)$ denote the predicted class. The flowchart in Figure~\ref{fig:OOD_flow} shows different sources of uncertainty that could make $\newinp$ an OOD.

\subsection{Additional experimental results}
We first present preliminary results for comparison with the OOD detection techniques based on fine-tuning of the classifiers~\citep{oe,energy,contrastive,self-supervised}.
We then present our results on various vision datasets and different architectures of the pre-trained  DNN based classifiers for these datasets in comparison to the ODIN, the Mahalanobis and the SPB methods in supervised settings. Finally, we then report results from the ablation study on OOD detection with individual attributes and compare it with our integrated approach on OOD detection.

\subsubsection{Comparison with the OOD detection techniques based on refinement of the training process for classifiers}
Recent techniques propose refinement in the training process of the classifiers for OOD detection. Some of these techniques include fine-tuning the training of classifiers with a trainable cost function for OOD detection~\citep{oe,energy}, self-supervised training of the classifiers to enhance OOD detection~\citep{contrastive,self-supervised} etc. 

We perform preliminary experiments to compare the performance of these techniques with our integrated OOD detector that uses features of the pre-trained classifiers to distinguish in-distribution samples from OODs. Table~\ref{table:comp-energy} compares TNR (at 95\% TPR), AUROC and AUPR for the energy based OOD detector~\citep{energy} and our integrated OOD detector on CIFAR10 with pre-trained WideResNet model. The integrated OOD detector was trained on in-distribution and adversarial samples generated by FGSM attack. Table~\ref{table:comp-oe} compares the results of the WideResNet model trained on CIFAR10 and fine-tuned with the outlier exposure from the 80 Million Tiny Images with our OOD detector that uses features from the pre-trained WideResNet model trained on CIFAR10. Since 80 Million Tiny Images dataset is no longer available for use, we used a small subset of ImageNet (treated as OOD dataset for CIFAR10 and SVHN datasets~\citep{mahalanobis}) for generating OODs for training of the integrated OOD detector. Table~\ref{table:comp-self-super} compares the OOD detection performance of the self-supervised training based OOD detector with our method. We trained our OOD detector with the in-distribution CIFAR10 as in-distribution samples and adversarial samples generated by FGSM attack from the test dataset of CIFAR10 as OODs. The trained OOD detector was then tested on LSUN as OODs and the results are reported in Table~\ref{table:comp-self-super}.  With ResNet-50 as the classifier for CIFAR10, we trained our OOD detector with the in-distribution CIFAR10 as in-distribution samples and adversarial samples generated by FGSM from the test dataset of CIFAR10 as OODs. The trained OOD detector was then tested on SVHN as OODs and these results are compared with the contrastive based learning for OOD detection~\citep{contrastive} in table~\ref{table:comp-contrastive}.

\begin{table}
\begin{center}
\caption{Results with Energy based OOD detector~\citep{energy} / Our method.}
\begin{tabular}{m{2cm} m{2cm} m{2cm} m{2cm} m{2cm}}
\hline \\
\multirow{2}{*}{} {\bf In-dist} & {\bf OOD} & {\bf TNR} & {\bf AUROC} & {\bf AUPR} \\
{\bf (model)} & {\bf dataset} & {\bf (TPR=95\%)}
\\
\\ \hline \\

\multirow{2}{*}{} CIFAR10 & SVHN            & 64.41 / \textbf{88.95}   &  90.96  / \textbf{97.61}  	&  97.64 / \textbf{92.84} 	\\

(WideResNet) & LSUN            & 72.42 / \textbf{98.84} 	 & 94.24 / \textbf{99.63} 	 &  98.67 /  \textbf{99.25}    \\

\\ \hline
\end{tabular}
\label{table:comp-energy}
\end{center}
\end{table}

\begin{table}
\begin{center}
\caption{Results with Outlier Exposure based OOD detector~\citep{oe} / Our method.}
\begin{tabular}{m{2cm} m{2cm} m{2cm} m{2cm} m{2cm}}
\hline \\
\multirow{2}{*}{} {\bf In-dist} & {\bf OOD} & {\bf TNR} & {\bf AUROC} & {\bf AUPR} \\
{\bf(model)} & {\bf dataset} & {\bf (TPR=95\%)}
\\
\\ \hline \\

\multirow{2}{*}{} CIFAR10 & SVHN            & \textbf{95.64} / 92.53  &   \textbf{98.63} / 98.56 	&  \textbf{99.74} / 96.62 	\\

(WideResNet)   \\

\\ \hline
\end{tabular}
\label{table:comp-oe}
\end{center}
\end{table}

\begin{table}
\begin{center}
\caption{Results with self-supervised learning based OOD detector~\citep{self-supervised} / Our method.}
\begin{tabular}{m{2cm} m{2cm} m{2cm} m{2cm} m{2cm}}
\hline \\
\multirow{2}{*}{} {\bf In-dist} & {\bf OOD} & {\bf TNR} & {\bf AUROC} & {\bf DTACC} \\
{\bf(model)} & {\bf dataset} & {\bf (TPR=95\%)}
 
\\
\\ \hline \\

\multirow{2}{*}{} CIFAR10 & LSUN            & 71.3 / \textbf{98.84}  &   93.2 / \textbf{99.63}  	&  71.0 / \textbf{97.72} 	\\

(WideResNet)   \\

\\ \hline
\end{tabular}
\label{table:comp-self-super}
\end{center}
\end{table}

\begin{table}
\begin{center}
\caption{Results with contrastive learning based OOD detector~\citep{contrastive} / Our method.}
\begin{tabular}{m{2cm} m{2cm} m{2cm} m{2cm} m{2cm} m{2cm}}
\hline \\
\multirow{2}{*}{} {\bf In-dist} & {\bf OOD} & {\bf TNR} & {\bf AUROC} & {\bf DTACC} & {\bf AUPR} \\
{\bf (model)} & {\bf dataset} & {\bf (TPR=95\%)}
 
\\
\\ \hline \\

\multirow{2}{*}{} CIFAR10 & SVHN           & \textbf{97.2} / 82.88   &   \textbf{99.5} /96.98 	&  \textbf{96.7} / 91.74 & \textbf{99.6} / 94.71	\\

(ResNet50)   \\

\\ \hline
\end{tabular}
\label{table:comp-contrastive}
\end{center}
\end{table}

\noop{The choice of these attributes makes the OOD detector's signature sound and complete in the following manner:
\begin{enumerate}
    \item \textbf{Sound:} Mahalanobis distance from the in-distribution density estimate tries to differentiate OODs due to epistemic uncertainty from the in-distributions samples. Reconstruction error from the class conditional PCA captures characteristics that differentiate in-distribution samples from the OODs due to epistemic uncertainty in class-wise distribution. Softmax scores will try to capture those OODs due to aleatory and epistemic uncertainty that lie close to the classifier boundary (or regions of low softmax scores). Conformance measure among the nearest neighbors tries to differentiate in-distribution samples for the OODs due to aleatory uncertainty.  
    \item \textbf{Complete:} OODs due to epistemic uncertainty in tied in-distribution tend to differ from the in-distribution samples in their distance from the in-distribution density estimate. Mahalanobis distance from the in-distribution density estimate tries to capture this difference. OODs due to epistemic uncertainty in the class-wise in-distribution tend to vary from the in-distribution in their density estimates from the class distributions. Reconstruction error from the class conditional PCA tends to capture this differentiating property. OODs due to the aleatory uncertainty tend to vary from the in-distribution samples in their conformance measure among the nearest neighbors. Mahalanobis distance of the conformance vector tries to capture this differentiating property between the OODs and in-distribution samples.
\end{enumerate}
}

\subsubsection{Comparison with the state-of-the-art OOD detection methods in supervised settings on pre-trained classifiers}
We compare our results with the state-of-the-art methods in supervised settings, as reported by the Mahalanobis method for OOD detection~\citep{mahalanobis}. In supervised settings, a small subset of the real OOD dataset is used in the training of the OOD detector. 

\textbf{Datasets and metrics.} 
We evaluate the proposed integrated OOD detection on  benchmarks such as MNIST~\citep{lenet}, CIFAR10~\citep{CIFAR10}, and SVHN~\citep{svhn}. We consider standard metrics~\citep{baseline,odin,mahalanobis} such as the true negative rate (TNR) at 95\% true positive rate (TPR), the area under the receiver operating characteristic curve (AUROC), area under precision recall curve (AUPR) with both in-distribution and OODs as positive samples (AUPR IN and AUPR OUT respectively), and the detection accuracy (DTACC) to evaluate our performance. 

\textbf{DNN-based classifier architectures.} To demonstrate that the proposed approach generalizes across various network architectures, we consider a wide range of DNN models such as Lenet~\citep{lenet} 
, ResNet~\citep{resnet} 
, and DenseNet~\citep{densenet}.

\textbf{Comparison with the state-of-the-art.} We compare our approach with the three state-of-the-art approaches:  SPB~\citep{baseline}, ODIN~\citep{odin} and Mahalanobis~\citep{mahalanobis}. Since, these experiments are performed in supervised settings, we fix $T=10$ and $\epsilon=0.005$ for generating results from the ODIN method. For Mahalanobis distance, we consider the distance in the penultimate layer feature space as well as features from all the layers of the DNN without preprocessing of the input in either settings.


\indent \textbf{MNIST.} With MNIST as in-distribution, we consider KMNIST~\citep{kmnist} and Fashion-MNIST(F-MNIST)~\citep{fashion-mnist} as OOD datasets. For MNIST, we use the LeNet5~\citep{lenet} DNN. Results in terms of TNR (at 95\% TPR), AUROC, and DTACC are reported in tables~\ref{table:penul-layer-results},~\ref{table:all-layers-results}, and~\ref{table:comp_baseline}. Table~\ref{table:penul-layer-results} shows the results with the features from the penultimate layer in comparison to the ODIN and Mahalanobis methods. Table~\ref{table:all-layers-results} shows the results with the features from all the layers in comparison to the Mahalanobis method. Table~\ref{table:comp_baseline} shows the results with the features from the penultimate layer in comparison to the SPB method. In all these settings, our approach outperforms the state-of-the-art approaches for both the OOD datasets. Results in comparison to AUPR IN and AUPR OUT are shown in table~\ref{table:mnist-aupr}. Here also, our technique out-performs all the three OOD detectors on all the test cases.

\indent \textbf{CIFAR10.} With CIFAR10 as in-distribution, we consider STL10~\citep{stl10}, SVHN~\citep{svhn}, Imagenet~\citep{imagenet}, LSUN~\citep{lsun}, and a subset of CIFAR100 (SCIFAR100)~\citep{CIFAR10} as OOD datasets. For CIFAR10, we consider three DNNs: DenseNet, ResNet34, and ResNet50. Results in terms of TNR (at 95\% TPR), AUROC, and DTACC are reported in tables~\ref{table:penul-layer-results},~\ref{table:all-layers-results}, and ~\ref{table:comp_baseline}. Table~\ref{table:penul-layer-results} shows the results with the features from the penultimate layer in comparison to the ODIN and Mahalanobis methods. Table~\ref{table:all-layers-results} shows the results with the features from all the layers in comparison to the Mahalanobis method. Table~\ref{table:comp_baseline} shows the results with the features from the penultimate layer in comparison to the SPB method. Results in comparison to AUPR IN and AUPR OUT are shown in tables~\ref{table:cifar10-densenet},~\ref{table:cifar10-resent34}, and~\ref{table:cifar10-resent50}. Here also, the integrated OOD detection technique could out-perform the other three detectors on most of the test cases. 

Note that images from STL10 and the subset of CIFAR100 are quite similar to CIFAR10 images. 
 Furthermore, from the CIFAR100 classes, we select sea, road, bee, and butterfly as OODs which are visually similar to the ship, automobile, and bird classes in the CIFAR10, respectively. 

\indent \textbf{SVHN.} With SVHN as in-distribution, we consider STL10, CIFAR10, Imagenet, LSUN and, SCIFAR100 as OOD datasets. For SVHN, we consider two DNNs: DenseNet and ResNet34. 
Results in terms of TNR (at 95\% TPR), AUROC, and DTACC are reported in tables~\ref{table:penul-layer-results},~\ref{table:all-layers-results}, and ~\ref{table:comp_baseline}. Table~\ref{table:penul-layer-results} shows the results with the features from the penultimate layer in comparison to the ODIN and Mahalanobis methods. Table~\ref{table:all-layers-results} shows the results with the features from all the layers in comparison to the Mahalanobis method. Table~\ref{table:comp_baseline} shows the results with the features from the penultimate layer in comparison to the SPB method. Results in comparison to AUPR IN and AUPR OUT are shown in tables~\ref{table:svhn-densenet}, and~\ref{table:svhn-resent34}. Here also, the integrated OOD detection technique could out-perform the other three detectors on most of the test cases. 

\indent \textbf{Key observations.} As shown in Table~\ref{table:penul-layer-results} and Table~\ref{table:all-layers-results}, and Table~\ref{table:comp_baseline}, our approach outperforms the state-of-the-art on all three datasets and with various DNN architectures. On CIFAR10, in terms of the TNR metric, our approach with Resnet50 outperforms Mahalanobis by 56\%  when SVHN is  OOD and our approach with Resnet34 outperforms ODIN by 36\% when LSUN is  OOD. 

While considering STL10 and Subset-CIFAR100 as OODs for CIFAR10, the images from both these datasets are quite similar to CIFAR-10 images. Thus, there can be numerous OOD samples due to aleatoric and class-conditional epistemic uncertainty which makes detection challenging. Although our performance is low on the STL10 dataset, it still outperforms the state-of-the-art. For instance, the proposed approach achieves a 27\% better TNR score than the Mahalanobis using ResNet50. On SVHN, in terms of the TNR metric, our approach outperforms ODIN and Mahalanobis by 63\% and 13\%, respectively on SCIFAR100 using ResNet34. The above observations justify the effectiveness of integrating multiple attributes to detect OOD samples.

\begin{table}
\begin{center}
\caption{Results with ODIN/Mahalanobis/Our method. The best results are highlighted.}
\begin{tabular}{m{1.5cm} m{1.6cm} m{2.9cm} m{2.9cm} m{2.9cm}}
\hline \\
\multirow{2}{*}{} {\bf In-dist} & {\bf OOD} & {\bf TNR} & {\bf AUROC} & {\bf DTACC} \\
{\bf (model)} & {\bf dataset} & {\bf (TPR=95\%)}
\\
\\ \hline \\

    \multirow{2}{*}{}  MNIST    & KMNIST   &       67.72 / 80.52 / \textbf{91.82}  	 &92.98 / 96.53 / \textbf{98.3}	&85.99 / 90.82 / \textbf{94.01}  \\
    (LeNet5)   & F-MNIST   &    58.47 / 63.33 / \textbf{74.49}  &90.76 / 94.11 / \textbf{95.55}	                           &83.21 / 87.76 / \textbf{90.98}	 \\
    
\\ \hline \\

    \multirow{5}{*}{} CIFAR10 & STL10  &  8.89 / 9.23 / \textbf{15.29} &56.31 / 62.16 / \textbf{63.96} &55.38 / 59.57 / \textbf{61.02}\\

(DenseNet) & SVHN   &   69.96 / 83.63 / \textbf{91.29}	&92.02 / 97.1 / \textbf{98.38}	&84.1 / 91.26 / \textbf{93.28}\\

& Imagenet  &   61.03 / 49.33 / \textbf{77.81}	&91.4 / 90.32 / \textbf{95.98}	&83.85 / 83.08 / \textbf{89.74}\\
          
& LSUN   &   71.89 / 46.63 / \textbf{84.34}	&94.37 / 91.18 / \textbf{97.27}	&87.72 / 84.93 / \textbf{92.1}\\
       
& SCIFAR100 &   35.06 / 20.33 / \textbf{38.78} &80.18 / 80.4 / \textbf{90.58}	&72.58 / 74.15 / \textbf{85.35}\\

\\ \hline \\

\multirow{5}{*}{} CIFAR10  & STL10   &10.63 / 13.9 / \textbf{17.4}	&61.56 / 66.47 / \textbf{67.52}	&59.22 / 62.75 / \textbf{63.7}	\\

(ResNet34) & SVHN    &72.85 / 53.16 / \textbf{88.2}	&93.85 / 93.85/  \textbf{97.69}	&85.4 / 89.173 / \textbf{92.14}	\\

& Imagenet  &46.54 / 68.41 / \textbf{74.53}	&90.45 / 95.02 / \textbf{95.73}	&83.06 / 88.63 / \textbf{89.73} \\

& LSUN    &45.16 / 77.53 / \textbf{81.23}	&89.63 / 96.51 / \textbf{96.87}	&81.83 / 90.64 / \textbf{91.19}	\\

& SCIFAR100   &37 / 38.39 / \textbf{61.11}	&86.13 / 88.86 / \textbf{94.74}	&78.5 / 82.51 / \textbf{90.53}\\

\\ \hline \\

\multirow{5}{*}{} CIFAR10 & STL10           &12.19 / 10.33 / \textbf{16}	&60.29 / 61.95 / \textbf{66.39}	&58.57 / 59.36 / \textbf{62.28}	\\

(ResNet50) & SVHN            &86.61 / 34.49/ \textbf{91.06} &84.41 / \textbf{98.19} / 91.98	&91.25 / 76.72 / \textbf{93.2}	\\

& Imagenet        &73.23 / 29.48 / \textbf{75.96}	&94.91 / 84.3 / \textbf{95.79}	&88.23 / 77.19 / \textbf{89.26}	\\

& LSUN            &80.72 / 32.18 / \textbf{81.38}	 &96.51 / 87.09 / \textbf{96.93}	 &90.59 / 80.07 / \textbf{91.79} \\

& SCIFAR100 &47.44 / 21.06 / \textbf{48.33}	&86.16 / 77.42/ \textbf{92.98}	&78.69 / 71.43 / \textbf{88.27} \\

\\ \hline \\

\multirow{5}{*}{} SVHN & STL10  &45.91 / 81.66 / \textbf{87.76}	 &77.6 / 96.97 / \textbf{97.63}	 &72.62 / 92.29 / \textbf{93.35}	 \\
       
(DenseNet) &CIFAR10   &37.23 / 80.82 / \textbf{86.42}	 &73.14 / 96.8 / \textbf{97.37}	 &68.92 / 92.27 / \textbf{92.86}	\\
       
& Imagenet  &62.76 / 85.44 / \textbf{93.44}	 &85.41 / 97.29 / \textbf{98.38}		 &79.94 / 93.39 / \textbf{94.53}\\
       
& LSUN   &62.91 / 76.87 / \textbf{89.73}	 &86.06 / 96.37 / \textbf{97.73}	 &80.04 / 92.43 / \textbf{93.55}\\
       
& SCIFAR100 &48.17 / 86.06 / \textbf{96.72}	 &78.94 / 97.43 / \textbf{98.24}	 &73.72 / 93.02 / \textbf{96.26}	\\
\\ \hline \\

\multirow{5}{*}{} SVHN & STL10  &35.14 / 85.3 / \textbf{90.9}	&67.05 / 97.19 / \textbf{97.76}	&66.19 / 93.41 / \textbf{94.34} \\

(ResNet34) & CIFAR10   &32.6 / 85.03 / \textbf{90.34}	&66.75 / 97.05 / \textbf{97.64} &65.37 / 93.15 / \textbf{94.29}	\\

& Imagenet &41.8 / 84.46 / \textbf{89.82}	&73 / 96.95 / \textbf{97.59}	&69.84 / 93.14 / \textbf{94.32}	\\

& LSUN   &35.92 / 78.38 / \textbf{85.46}	&68.6 / 96.17 / \textbf{97.09}	&66.75 / 91.98 / \textbf{93.17} \\

& SCIFAR100 &36.67 / 86.61 / \textbf{99.61}	&68.01 / 97.3 / \textbf{98.47}	&67.26 / 93.6 / \textbf{97.36} \\

\\ \hline
\end{tabular}
\label{table:penul-layer-results}
\end{center}
\end{table}

\begin{table}
\begin{center}
\caption{Results with Mahalanobis/Our method with feature ensemble. The best results are highlighted.}
\begin{tabular}{m{1.5cm} m{1.6cm} m{2cm} m{2cm} m{2cm}}
\hline \\
\multirow{2}{*}{} {\bf In-dist} & {\bf OOD} & {\bf TNR} & {\bf AUROC} & {\bf DTACC} \\
{\bf (model)} & {\bf dataset} & {\bf (TPR=95\%)}
\\
\\ \hline \\

    \multirow{2}{*}{}  MNIST    & KMNIST   &     96 / \textbf{98.8}	&99.19 / \textbf{99.65}	&95.56 / \textbf{97.3}	\\
    (LeNet5)   & F-MNIST   &   99.9 / \textbf{99.98}	 &99.95 / \textbf{99.96} &98.98 / \textbf{99.17} 	 \\
    
\\ \hline \\

    \multirow{5}{*}{} CIFAR10 & STL10  &16.44 / \textbf{22.94}	&72.4 / \textbf{75.23}	&66.69 / \textbf{69.31}\\

(DenseNet) & SVHN   &92.4 / \textbf{98.23}	&98.41 / \textbf{99.49}	&93.97 / \textbf{97.02}\\

& Imagenet     &96.46 / \textbf{98.8}	&99.16/ \textbf{99.63}	&95.74 / \textbf{97.55}\\
          
& LSUN  &98.09 / \textbf{99.64}	&99.47 / \textbf{99.85}	&96.76 / \textbf{98.45}\\
       
& SCIFAR100 &27.33 / \textbf{46.5}	&83.7 / \textbf{92.17}	&77.08 / \textbf{86.69}\\

\\ \hline \\

\multirow{5}{*}{} CIFAR10  & STL10   &26.14 / \textbf{29.8}	&76.23 / \textbf{76.46}	&70.33 / \textbf{70.94}	\\

(ResNet34) & SVHN    &91.53 / \textbf{97.07}	&98.4 / \textbf{99.32}	&93.63 / \textbf{96.27}	\\

& Imagenet &97.09 / \textbf{98.11}	&99.47 / \textbf{99.58}	&96.31 / \textbf{96.91} \\

& LSUN    &98.67 / \textbf{99.41}	 &99.71 / \textbf{99.81}	 &97.56 / \textbf{98.14}	\\

& SCIFAR100  &38.89 / \textbf{62.78}	&88.8 / \textbf{94.23}	&82.14 / \textbf{90.16}\\

\\ \hline \\

\multirow{5}{*}{} CIFAR10 & STL10  &26.36 / \textbf{30.83}	&73.74 / \textbf{76.73}	&67.37 / \textbf{70.4}	\\

(ResNet50) & SVHN   &84.44 / \textbf{98.59}	&96.56 / \textbf{99.65}	&90.63 / \textbf{97.43}\\

& Imagenet        &97.87 / \textbf{99.46}	&99.58 / \textbf{99.84}	&97.09 / \textbf{98.22}	\\

& LSUN            &99.21 / \textbf{99.83}	&99.64 / \textbf{99.91}	&\textbf{98.39} /99.21\\

& SCIFAR100 &29.33 / \textbf{55}	&80.26 / \textbf{91.48}	&74.51 / \textbf{86.42} \\

\\ \hline \\

\multirow{5}{*}{} SVHN & STL10  &97.31 / \textbf{98.76}	&99.14 / \textbf{99.47}	&96.23 / \textbf{97.24}	 \\
       
(DenseNet) &CIFAR10   &96.36 / \textbf{97.64}	&98.8 / \textbf{99.16}	&95.7 / \textbf{96.34}	\\
       
& Imagenet  &\textbf{99.89} / 99.82	&99.88 / \textbf{99.9}	&98.85 / \textbf{98.95}\\
       
& LSUN   &\textbf{99.99} / 99.97	&\textbf{99.91} / \textbf{99.91}	&\textbf{99.26} /99.18\\
       
& SCIFAR100 &99.33 / \textbf{100}	&99.53 / \textbf{99.78}	&97.89 / \textbf{98.95}	\\
\\ \hline \\

\multirow{5}{*}{} SVHN & STL10 &98.44 / \textbf{98.88}	&99.31 / \textbf{99.52}	&96.91 / \textbf{97.4} \\

(ResNet34) & CIFAR10   &98.44 / \textbf{98.88}	&99.31 / \textbf{99.52}	&96.91 / \textbf{97.4}	\\

& Imagenet &99.83 / \textbf{99.87}	&99.85 / \textbf{99.91}	&\textbf{99.07} / \textbf{99.07}		\\

& LSUN  &99.87 / \textbf{99.99}	&99.83 / \textbf{99.95}	&99.5 / \textbf{99.47} \\

& SCIFAR100 &99.83 / \textbf{100}	&99.72 / \textbf{99.91}	&98.33/ \textbf{99.56}\\

\\ \hline

\end{tabular}
\label{table:all-layers-results}
\end{center}
\end{table}

\begin{table}
\begin{center}
\caption{Experimental results with SPB/Our method. The best results are highlighted.}
\label{penul-layer-results}
\begin{tabular}{lllll}
\hline \\
\multirow{2}{*}{} {\bf In-dist} & {\bf OOD} & {\bf TNR} & {\bf AUROC} & {\bf DTACC} \\
{\bf (model)} & {\bf dataset} & {\bf (TPR=95\%)}
\\
\\ \hline \\

    \multirow{2}{*}{}  MNIST    & KMNIST   &       69.33 / \textbf{91.82}  	 &93.24 / \textbf{98.3}	&86.88 / \textbf{94.01}  \\
    (LeNet5)   & F-MNIST   &    52.69 / \textbf{74.49}  &89.19 / \textbf{95.55}	                           &82.77 / \textbf{90.98}	 \\
    
\\ \hline \\

    \multirow{5}{*}{} CIFAR10 & STL10  &  \textbf{15.64} / 15.29 &\textbf{64.15} / 63.96 &\textbf{62.12} / 61.02\\

(DenseNet) & SVHN   &   39.22 / \textbf{91.29}	&88.24 / \textbf{98.38}	&82.41 / \textbf{93.28}\\

& Imagenet  &   40.13 / \textbf{77.81}	&89.3 / \textbf{95.98}	&82.67 / \textbf{89.74}\\
          
& LSUN   &   48.38 / \textbf{84.34}	&92.14 / \textbf{97.27}	&86.22 / \textbf{92.1}\\
       
& SCIFAR100 &   34.11 / \textbf{38.78} & 85.53 / \textbf{90.58}	&79.18 / \textbf{85.35}\\

\\ \hline \\

\multirow{5}{*}{} CIFAR10  & STL10   &14.9 / \textbf{17.4}	&65.88 / \textbf{67.52}	&62.85 / \textbf{63.7}	\\

(ResNet34) & SVHN    &32.47 / \textbf{88.2}	&89.88 / \textbf{97.69}	&85.06 / \textbf{92.14}	\\

& Imagenet  &44.72 / \textbf{74.53}	&91.02 / \textbf{95.73}	&85.05 / \textbf{89.73} \\

& LSUN    &45.44 / \textbf{81.23}	&91.04 / \textbf{96.87}	&85.26 / \textbf{91.19}	\\

& SCIFAR100   &38.17 / \textbf{61.11}	&88.91 / \textbf{94.74}	&82.34 / \textbf{90.53}\\

\\ \hline \\

\multirow{5}{*}{} CIFAR10 & STL10  &15.33 / \textbf{16}	&66.68 / \textbf{66.39}	&63.47 / \textbf{62.28}	\\

(ResNet50) & SVHN            &44.69 / \textbf{91.06} &97.31 / \textbf{91.98}	&86.36 /\textbf{93.2}	\\

& Imagenet        &42.06 / \textbf{75.96}	&90.8 /\textbf{95.79}	&84.36 / \textbf{89.26}	\\

& LSUN            &48.37 / \textbf{81.38}	 &92.78 / \textbf{96.93}	 &86.97 / \textbf{91.79} \\

& SCIFAR100 &36.39 / \textbf{48.33}	&89.09 / \textbf{92.98}	&83.37 / \textbf{88.27} \\

\\ \hline \\

\multirow{5}{*}{} SVHN & STL10  &72.87 / \textbf{87.76}	 &92.79 / \textbf{97.63}	 &87.76 / \textbf{93.35}	 \\
       
(DenseNet) &CIFAR10   &69.31 / \textbf{86.42}	 &91.9 / \textbf{97.37}	 &86.61 / \textbf{92.86}	\\
       
& Imagenet  &79.79 / \textbf{93.44}	 &94.78 / \textbf{98.38}		 &90.21 / \textbf{94.53}\\
       
& LSUN   &77.12 / \textbf{89.73}	 &94.13 / \textbf{97.73}	 &89.14 / \textbf{93.55}\\
       
& SCIFAR100 &76.94 / \textbf{96.72}	 &94.18 / \textbf{98.24}	 &89.57 / \textbf{96.26}	\\
\\ \hline \\

\multirow{5}{*}{} SVHN & STL10  &79.57 / \textbf{99.59}	&93.84 / \textbf{99.72}	&90.83 / \textbf{98.06} \\

(ResNet34) & CIFAR10   &78.26 / \textbf{90.34}	&92.92 / \textbf{97.64} &90.03 / \textbf{94.29}	\\

& Imagenet &79.02 / \textbf{89.82}	&93.51 / \textbf{97.59}	&90.44 / \textbf{94.32}	\\

& LSUN   &74.29 / \textbf{85.46}	&91.58 / \textbf{97.09}	&88.96 / \textbf{93.17} \\

& SCIFAR100 &81.28 / \textbf{99.61}	&94.62 /\textbf{98.47}	&91.48 / \textbf{97.36} \\

\\ \hline

\label{table:comp_baseline}

\end{tabular}
\end{center}
\end{table}


\begin{table}
\caption{Experimental Results with MNIST on Lenet5 for AUPR IN and AUPR OUT. \\ The best results are highlighted.}
\begin{center}
\begin{tabular}{llllllll}
\multicolumn{1}{c}{\bf OOD Dataset} &\multicolumn{1}{c}{\bf Layer}  &\multicolumn{1}{c}{\bf Method} &\multicolumn{1}{c}{\bf AUPR IN} &\multicolumn{1}{c}{\bf AUPR OUT}
\\ \hline \\
         & Penultimate 
\\ \hline \\
KMNIST    &   &Baseline          	&92.47	&92.41 \\
          &   &ODIN             	&92.65	&92.69\\
          &   &Mahalanobis       	&96.69	&96.2 \\
          &   &Ours                 &\textbf{98.48}	&\textbf{98.13}\\
\\ \hline \\
Fashion-MNIST   &   &Baseline       &87.98	 &87.89\\
                &   &ODIN           &90.94	 &89.99\\
                &   &Mahalanobis  	&95.24	 &91.94\\
                &   &Ours           &\textbf{96.64}	&\textbf{92.96}\\
\\ \hline \\
         & All
\\ \hline \\
KMNIST 
          &   &Mahalanobis       	&99.22	&99.18 \\
          &   &Ours              	&\textbf{99.67}	&\textbf{99.64}\\
\\ \hline \\
Fashion-MNIST 
                &   &Mahalanobis   	&99.95	 &99.94\\
                &   &Ours          	&\textbf{99.96}	&\textbf{99.96}\\
\\ \hline 

\end{tabular}
\end{center}
\label{table:mnist-aupr}
\end{table}

\begin{table}
\caption{Experimental Results with CIFAR10 on DenseNet for AUPR IN and AUPR OUT. \\ The best results are highlighted.}
\begin{center}
\begin{tabular}{llllllll}
\multicolumn{1}{c}{\bf OOD Dataset} &\multicolumn{1}{c}{\bf Layer}  &\multicolumn{1}{c}{\bf Method} &\multicolumn{1}{c}{\bf AUPR IN} &\multicolumn{1}{c}{\bf AUPR OUT}
\\ \hline \\
 &Penultimate
\\ \hline \\
STL10  &   &Baseline        &64.55	&\textbf{59.37} \\
       &   &ODIN  	        &60.24	&50.17\\
       &   &Mahalanobis 	&65.86	&53.87\\
       &   &Ours            &\textbf{66.01}	&58.5\\
       \\ \hline \\
SVHN   &   &Baseline        &74.53	&94.09\\
       &   &ODIN            &80.49	&97.05\\
       &   &Mahalanobis 	&94.13	&98.78\\
       &   &Ours           &\textbf{96.26}	&\textbf{99.37}\\
       \\ \hline \\
Imagenet  &   &Baseline &90.88	&86.74\\
          &   &ODIN  	&91.32	&90.55\\
          &   &Mahalanobis &91.32	&88.6\\
          &   &Ours 	&\textbf{96.22}	&\textbf{95.68}\\
          \\ \hline \\
LSUN   &   &Baseline 	&93.68	&89.83\\
       &   &ODIN     	&94.65 &93.39\\
       &   &Mahalanobis 	&92.71 &87.74\\
       &   &Ours 	&\textbf{97.6}	&\textbf{96.74}\\
       \\ \hline \\
Subset CIFAR100 &   &Baseline 	&96.65	&50.08 \\
                &   &ODIN 	&95.14	&47.64\\
                &   &Mahalanobis 	&95.68	&37.86 \\
                &   &Ours &\textbf{98.18}	&\textbf{54.83}\\
\\ \hline \\
&All
\\ \hline \\
STL10 
       &   &Mahalanobis 	&77.21	&63.45\\
       &   &Ours 	&\textbf{78.29}	&\textbf{68.14}\\
       \\ \hline \\
SVHN 
       &   &Mahalanobis 	&96.72	&99.31\\
       &   &Ours            &\textbf{98.57}	&\textbf{99.81}\\
       \\ \hline \\
Imagenet
       &   &Mahalanobis 	&99.19	&99.13\\
       &   &Ours 	&\textbf{99.62}	&\textbf{99.54} \\
       \\ \hline \\
LSUN 
       &   &Mahalanobis 	&99.49	&99.45 \\
       &   &Ours 	&\textbf{99.82}	&\textbf{99.85}\\
       \\ \hline \\
Subset CIFAR100 
       &   &Mahalanobis 	&96.58	&42.41\\
       &   &Ours    &\textbf{98.53}	&\textbf{59.7}
\\ \hline \\
\end{tabular}
\end{center}
\label{table:cifar10-densenet}
\end{table}

\begin{table}
\caption{Experimental Results with CIFAR10 on ResNet34 for AUPR IN and AUPR OUT. \\ The best results are highlighted.}
\begin{center}
\begin{tabular}{ccccc}
\multicolumn{1}{c}{\bf OOD Dataset} &\multicolumn{1}{c}{\bf Layer}  &\multicolumn{1}{c}{\bf Method}  &\multicolumn{1}{c}{\bf AUIN} &\multicolumn{1}{c}{\bf AUOUT}
\\ \hline \\
&Penultimate
\\ \hline \\
STL10  &   &Baseline	&67.17	&59.74\\
       &   &ODIN 	&64.22	&53.83\\
       &   &Mahalanobis 	&68.48	&59.47\\
       &   & Ours	&\textbf{68.78}	&\textbf{61.52}\\
 \\ \hline \\   
SVHN   &   &Baseline 	&85.4	&93.96\\
       &   &ODIN  	&86.46	&97.55\\
       &   &Mahalanobis	&91.19	&96.14\\
       &   & Ours  	&\textbf{94.7}	&\textbf{99.1}\\
 \\ \hline \\ 
Imagenet &   &Baseline  	&92.49	&88.4 \\
       &   &ODIN    	&92.11	&87.46\\
       &   &Mahalanobis    	&95.77	&94.02\\
       &   & Ours   	&\textbf{96.32}	&\textbf{94.99}\\
        \\ \hline \\ 
LSUN   &   &Baseline	&92.45	&88.55\\
       &   &ODIN     	&91.58	&86.5\\
       &   &Mahalanobis 	&97.08	&95.78\\
       &   & Ours  &\textbf{97.36}	&\textbf{96.29}\\
        \\ \hline \\ 
Subset CIFAR100 &   &Baseline   	&97.77	&55.62\\
                &   &ODIN  	&97.05	&51.57\\
                &   &Mahalanobis 	&97.71	&54.11\\
                &   & Ours 	&\textbf{99.06}	&\textbf{64.53}\\
\\ \hline \\
&All
\\ \hline \\
STL10 
       &   &Mahalanobis	&77.59	&71.15\\
       &   & Ours  	&\textbf{77.32}	&\textbf{72.38}\\
       \\ \hline \\   
SVHN  
       &   &Mahalanobis 	&96.46	&99.37\\
       &   & Ours  &\textbf{98.37}	&\textbf{99.73}\\
       \\ \hline \\   
Imagenet
       &   &Mahalanobis 	 &99.48	 &99.48\\
       &   & Ours   &\textbf{99.59}	&\textbf{99.58}\\
       \\ \hline \\   
LSUN   
       &   &Mahalanobis   &99.71	 &99.71\\
       &   & Ours  &\textbf{99.8}	&\textbf{99.82}\\
       \\ \hline \\   
Subset CIFAR100 
                &   &Mahalanobis 	&97.75	&52.28\\
                &   & Ours  	&\textbf{98.74}	&\textbf{65.99}\\ 
\\ \hline 
\end{tabular}
\end{center}
\label{table:cifar10-resent34}
\end{table}

\begin{table}
\caption{Experimental Results with CIFAR10 as on ResNet50 for AUPR IN and AUPR OUT. \\ The best results are highlighted.}
\begin{center}
\begin{tabular}{ccccc}
\multicolumn{1}{c}{\bf OOD Dataset} &\multicolumn{1}{c}{\bf Layer}  &\multicolumn{1}{c}{\bf Method} &\multicolumn{1}{c}{\bf AUIN} &\multicolumn{1}{c}{\bf AUOUT}
\\ \hline \\
&Penultimate
\\ \hline \\
STL10           &   &Baseline       	&67.47	&60.83\\
                &   &ODIN           	&62.79	&55.04 \\
                &   &Mahalanobis    	&65.14	&54.43\\
                &   & Ours         	&\textbf{68.54}	&\textbf{59.75}\\
                \\ \hline \\
SVHN            &   &Baseline    	&87.78	&95.61 \\
                &   &ODIN        	&93.17	&99.03 \\
                &   &Mahalanobis 	&71.88	&92.54   \\
                &   & Ours       	&\textbf{95.38}	&\textbf{99.34}\\
                \\ \hline \\
Imagenet        &   &Baseline    	&92.6	&87.98 \\
                &   &ODIN        	&95.16	&94.45\\
                &   &Mahalanobis 	&86.14	&80.6   \\
                &   & Ours      	&\textbf{96.22}	&\textbf{95.23} \\
                                \\ \hline \\
LSUN            &   &Baseline     &94.45	 &90.41 \\
                &   &ODIN         	&96.9	 &96.01\\
                &   &Mahalanobis   	&89.34	 &82.87  \\
                &   & Ours         	 &\textbf{97.53}	 &\textbf{96.03}\\
                                \\ \hline \\
Subset CIFAR100 &   &Baseline     	&97.72	&55.29 \\
                &   &ODIN         	&96.67	&60.62 \\
                &   &Mahalanobis  	&94.49	&36.12  \\
                &   & Ours        	&\textbf{98.72}	&\textbf{59.3}\\
\\ \hline \\
&All
\\ \hline \\
STL10 
       &   &Mahalanobis    	&75.6	&69.32\\
       &   & Ours         	&\textbf{77.79}	&\textbf{73.22}\\
                       \\ \hline \\
SVHN   
       &   &Mahalanobis 	&91.89	&98.58\\
       &   & Ours     	&\textbf{99.12}	&\textbf{99.86}\\
                       \\ \hline \\
Imagenet
         &   &Mahalanobis 	&99.56	&99.6\\
         &   & Ours       	&\textbf{99.84}	&\textbf{99.84}\\
                         \\ \hline \\
LSUN  
       &   &Mahalanobis  &\textbf{98.91}	&99.75\\
       &   & Ours        	&99.72	&\textbf{99.93}\\
                       \\ \hline \\
Subset CIFAR100
                &   &Mahalanobis  	&94.1	&42.96\\
                &   & Ours        	&\textbf{97.54}	&\textbf{64.12}\\ 
\\ \hline 
\end{tabular}
\end{center}
\label{table:cifar10-resent50}
\end{table}

\begin{table}
\caption{Experimental Results with SVHN as on DenseNet for AUPR IN and AUPR OUT. \\ The best results are highlighted.}
\label{sample-table5}
\begin{center}
\begin{tabular}{ccccc}
\multicolumn{1}{c}{\bf OOD Dataset} &\multicolumn{1}{c}{\bf Layer}  &\multicolumn{1}{c}{\bf Method} &\multicolumn{1}{c}{\bf AUIN} &\multicolumn{1}{c}{\bf AUOUT}
\\ \hline \\
&Penultimate
\\ \hline \\
STL10  &   &Baseline  	 &97.01	 &82.02  \\
       &   &ODIN    	 &89.18	 &63.71\\
       &   &Mahalanobis   &99.2	 &87.32\\
       &   & Ours   	&\textbf{99.36}	&\textbf{90.31}\\
              \\ \hline \\

CIFAR10   &   &Baseline   &95.7	 &82.8   \\
       &   &ODIN      &84.32 &	60.32\\
       &   &Mahalanobis  	 &98.94	 &88.91\\
       &   & Ours	&\textbf{99.09}	&\textbf{91.59}\\
       \\ \hline \\
Imagenet &   &Baseline    &97.2	 &88.42\\
       &   &ODIN     &90.95	 &79.59\\
       &   &Mahalanobis  	 &99.12	 &90.22\\
       &   & Ours  	&\textbf{99.4}	&\textbf{95.15}\\
       \\ \hline \\
LSUN   &   &Baseline      &96.96	 &87.44\\
       &   &ODIN      &92.03	 &79.98\\
       &   &Mahalanobis      &98.84	 &85.79\\
       &   & Ours   &\textbf{99.17}	&\textbf{92.92}\\
       \\ \hline \\
Subset CIFAR100 &   &Baseline   	 &99.39	 &63.21\\
                &   &ODIN     &97.24	 &45.23\\
                &   &Mahalanobis     	 &99.82	 &\textbf{72.35}\\
                &   & Ours   	&\textbf{99.88}	&68.25\\
\\ \hline \\
&All
\\ \hline \\
STL10 
       &   &Mahalanobis 	&99.75	&96.51\\
       &   & Ours   	&\textbf{99.77}	&\textbf{98.18}\\
              \\ \hline \\

CIFAR10   
       &   &Mahalanobis 	&99.6	&95.39\\
       &   & Ours  	&\textbf{99.69}	&\textbf{97.21}\\
              \\ \hline \\

Imagenet 
       &   &Mahalanobis	&\textbf{99.96}	&99.59\\
       &   & Ours   	&\textbf{99.96}	&\textbf{99.74}\\
              \\ \hline \\

LSUN   
       &   &Mahalanobis	&\textbf{99.97}	&99.7\\
       &   & Ours   	&99.95  &\textbf{99.74}\\
              \\ \hline \\

Subset CIFAR100 
                &   &Mahalanobis   	&99.97	&91.41\\
                &   & Ours 	&\textbf{99.98}	&\textbf{94.54}\\ 
\\ \hline 
\end{tabular}
\end{center}
\label{table:svhn-densenet}
\end{table}

\begin{table}
\caption{Experimental Results with SVHN as on ResNet34 for AUPR IN and AUPR OUT. \\ The best results are highlighted.}
\label{sample-table6}
\begin{center}
\begin{tabular}{ccccc}
\multicolumn{1}{c}{\bf OOD Dataset} &\multicolumn{1}{c}{\bf Layer}  &\multicolumn{1}{c}{\bf Method} &\multicolumn{1}{c}{\bf AUIN} &\multicolumn{1}{c}{\bf AUOUT}
\\ \hline \\
&Penultimate
\\ \hline \\
STL10  &   &Baseline    	&96.63	&84.15\\
       &   &ODIN        	&84.03	&47.26\\
       &   &Mahalanobis 	&99.89	&97.5\\
       &   & Ours       	&\textbf{99.93}	&\textbf{98.2}\\
       \\ \hline \\
CIFAR10   &   &Baseline 	&95.06	&85.66\\
       &   &ODIN        	&80.69	&50.49\\
       &   &Mahalanobis 	&99.04	&88.62\\
       &   & Ours       &\textbf{99.17}	&\textbf{91.17}\\
       \\ \hline \\
Imagenet &   &Baseline  &95.68	&86.18\\
       &   &ODIN        &84.62	&58.28\\
       &   &Mahalanobis &99	&88.39\\
       &   & Ours      &\textbf{99.19}	&\textbf{90.77}\\
       \\ \hline \\
LSUN   &   &Baseline    &94.19	&83.95\\
       &   &ODIN        &82.37	&53.12\\
       &   &Mahalanobis &98.73	&85.11\\
       &   & Ours       &\textbf{99.03}	&\textbf{89.03}\\
       \\ \hline \\
Subset CIFAR100 &       &Baseline    &99.35	&64.38\\
       &   &ODIN        &95.57	&23.04\\
       &   &Mahalanobis &99.81	&64.4\\
       &   & Ours      &\textbf{99.89}	&\textbf{67.9}
\\ \hline \\
&All
\\ \hline \\
STL10  
       &   &Mahalanobis 	&99.7	&97.03\\
       &   & Ours      	&\textbf{99.84}	&\textbf{97.86}\\
       \\ \hline \\
CIFAR10   
       &   &Mahalanobis 	&99.7	&97.03\\
       &   & Ours       	&\textbf{99.84}	&\textbf{97.86}\\
       \\ \hline \\
Imagenet 
       &   &Mahalanobis 	&99.86	&99.14   \\
       &   & Ours       	&\textbf{99.93}	&\textbf{99.62}\\
       \\ \hline \\
LSUN   
       &   &Mahalanobis &99.82	&98.85\\
       &   & Ours       	&\textbf{99.98}	&\textbf{99.64}\\
       \\ \hline \\
Subset CIFAR100 
                &   &Mahalanobis 	&99.98	&93.59\\
                &   & Ours       	&\textbf{99.99}	&\textbf{95.75}\\ 
\\ \hline 
\end{tabular}
\end{center}
\label{table:svhn-resent34}
\end{table}

\subsubsection{Ablation study}
We report ablation study on OOD detection with individual attributes and compare it with our integrated approach on the penultimate feature space of the classifier in the supervised settings as described in the previous section. We call the OOD detector with Mahalanobis distance estimated on class mean and tied covariance~\citep{mahalanobis} as Mahala-Tied. Detector based on Mahalanobis distance estimated on class mean and class covariance is referred as Mahala-Class. Similarly conformance among the K-nearest neighbors (KNN) measured by Mahala-Tied and Mahala-Class is referred as KNN-Tied and KNN-Class respectively in these experiments. Results for this study on CIFAR10 with DenseNet architecture, SVHN with DenseNet and ResNet34 architectures are shown in Tables~\ref{table:ablation-cifar},~\ref{table:ablation-svhn-densenet} and~\ref{table:ablation-svhn} respectively. 

The integrated approach could out-perform all the single attribute based OOD detector in all the tested cases due to detection of diverse OODs. An important observation made from these experiments is that the performance of the single attribute based methods could depend on the architecture of the classifier.  For example, while the performance of PCA was really bad in case of DenseNet (for both CIFAR10 and SVHN) as compared to all other methods, it could out-perform all but the integrated approach for SVHN on ResNet34.

\begin{table}

\caption{Ablation study with CIFAR10 on DenseNet. \\ The best results are highlighted.}
\begin{center}
\begin{tabular}{ccccccc}
\label{table:ablation-cifar}
\\ \hline \\
\multirow{2}{*}{}  {\bf OOD} & {\bf Method} & {\bf TNR} & {\bf AUROC} & {\bf DTACC} & {\bf AUPR} & {\bf AUPR} \\
 {\bf dataset} &  & {\bf (TPR=95\%)} &  & &{\bf IN} &{\bf OUT}\\
\\ \hline \\

SVHN   &   Mahala-Tied        &83.63	&97.1	&91.26	&94.13	&98.78\\
       &   Mahala-Class	      &71.73	&95.16	&87.92	&90.77	&97.98\\
       &   KNN-Tied 	      &84.07	&97.18	&91.32	&94.2	&98.84\\
       &   KNN-Class 	      &77.95	&96.19	&89.68	&92.06	&98.45\\
       &   SPB	              &39.22	&88.24	&82.41	&74.53	&94.09\\
       &   ODIN	              &69.96	&92.02	&84.1	&80.49	&97.05\\
       &   PCA                &2.46	    &55.89	&56.36	&35.42	&74.12 \\
       &   Integrated(Our)         &\textbf{91.29}	&\textbf{98.38}	&\textbf{93.28}	&\textbf{96.26}	&\textbf{99.37}\\
       \\ \hline \\
Imagenet &   Mahala-Tied &49.33	&90.32	&83.08	&91.32	&88.6\\
       &   Mahala-Class	&53.11	&92.16	&85.3	&93.42	&90.29\\
       &   KNN-Tied     &51.36	&90.73	&83.31	&91.75	&88.87\\
       &   KNN-Class    &57.94  &92.74	&86.01	&93.67	&91.28\\
       &   SPB	        &40.13	&89.3	&82.67	&90.88	&86.74\\
       &   ODIN	        &61.03	&91.4	&83.85	&91.32	&90.55\\
       &   PCA          &4.66	&58.68	&57.19	&60.66	&54.42 \\
       &   Integrated(Our)           &\textbf{77.81}	&\textbf{95.98}	&\textbf{89.74}	&\textbf{96.22}	&\textbf{95.68}\\
          \\ \hline \\
LSUN   &   Mahala-Tied  &46.63	&91.18	&84.93	&92.71	&87.74\\
       &   Mahala-Class	&58.53	&93.82	&88.16	&95.15	&91.43\\
       &   KNN-Tied 	&51.48	&92.25	&85.96	&93.75	&89.13\\
       &   KNN-Class 	&65.17	&94.57	&88.6	&95.61	&92.61\\
       &   SPB	        &48.38	&92.14	&86.22	&93.68	&89.83\\
       &   ODIN	        &71.89	&94.37	&87.72	&94.65	&93.39\\
       &   PCA          &2.06	&53.26	&54.88	&57.08	&49.33\\
       &   Integrated(Our)           &\textbf{84.34}	&\textbf{97.27}	&\textbf{92.1}	&\textbf{97.6}	&\textbf{96.74}\\
       \\ \hline \\
\end{tabular}
\end{center}
\label{table:cifar10-densenet1}
\end{table}

\begin{table}
\caption{Ablation study with SVHN on DenseNet. \\ The best results are highlighted.}
\begin{center}
\begin{tabular}{ccccccc}
\hline \\
\multirow{2}{*}{}  {\bf OOD} & {\bf Method} & {\bf TNR} & {\bf AUROC} & {\bf DTACC} & {\bf AUPR} & {\bf AUPR} \\
 {\bf dataset} &  & {\bf (TPR=95\%) }&  & &{\bf IN} &{\bf OUT}\\
\\ \hline \\

CIFAR10   &   Mahala-Tied     &80.82	&96.8	&92.27	&98.94	&88.91\\
       &   Mahala-Class	      &82.99	&97.11	&92.83	&99.05	&89.71\\
       &   KNN-Tied 	      &69.99	&95.58	&90.77	&98.52	&84.3\\
       &   KNN-Class 	      &74.52	&96.01	&91.21	&98.64	&85.99\\
       &   SPB	              &69.31	&91.9	&86.61	&95.7	&82.8\\
       &   ODIN	              &37.23	&73.14	&68.92	&84.32	&60.32\\
       &   PCA                &5.27	&65.82	&64.83	&86.62	&33.51 \\
       &   Integrated(Our)         &\textbf{86.6}	&\textbf{97.41}	&\textbf{92.88}	&\textbf{99.11}	&\textbf{91.76}\\
       \\ \hline \\
Imagenet   &   Mahala-Tied &85.44	&97.29	&93.39	&99.12	&90.22\\
       &   Mahala-Class	  &77.66	&96.83	&93.17	&98.98	&88.59\\
       &   KNN-Tied 	  &65.76	&94.67	&89.59	&98.18	&80.16\\
       &   KNN-Class 	  &73.44	&95.69	&90.68	&98.55	&84.28\\
       &   SPB	          &79.79	&94.78	&90.21	&97.2	&88.42\\
       &   ODIN	          &62.76	&85.41	&79.94	&90.95	&79.59\\
       &   PCA            &5.16 	&65.08	&65.39	&86.65	&32.83\\
       &   Integrated(Our)           &\textbf{93.46}	&\textbf{98.39}	&\textbf{94.54}	&\textbf{99.41}	&\textbf{95.16}\\
       \\ \hline \\
LSUN &   Mahala-Tied    &76.87	&96.37	&92.43	&98.84	&85.79\\
       &   Mahala-Class	&69.44	&96.05	&92.4	&98.74	&84.89\\
       &   KNN-Tied     &59.64	&93.71	&88.22	&97.83	&77.17\\
       &   KNN-Class    &66.96	&94.77	&89.45	&98.21	&81.27\\
       &   SPB	        &77.12	&94.13	&89.14	&96.96	&87.44\\
       &   ODIN	        &62.91	&86.06	&80.04	&92.03	&79.98\\
       &   PCA          &3.19	&62.66	&64.7	&85.72	&30.37 \\
       &   Integrated(Our)           &\textbf{89.73}	&\textbf{97.73}	&\textbf{93.55}	&\textbf{99.17}	&\textbf{92.92}\\
          \\ \hline \\
\end{tabular}
\end{center}
\label{table:ablation-svhn-densenet}
\end{table}

\begin{table}
\caption{Ablation study with SVHN on ResNet34. \\ The best results are highlighted.}
\begin{center}
\begin{tabular}{ccccccc}
\hline \\
\multirow{2}{*}{}  {\bf OOD} & {\bf Method} & {\bf TNR} & {\bf AUROC} & {\bf DTACC} & {\bf AUPR} & {\bf AUPR} \\
 {\bf dataset} &  & {\bf (TPR=95\%)} &  & &{\bf IN} &{\bf OUT}\\
\\ \hline \\

SCIFAR100   &   Mahala-Tied   &86.61	&97.3	&93.6	&99.81	&64.4\\
       &   Mahala-Class	      &88.44	&97.7	&94.19	&99.84	&69.59\\
       &   KNN-Tied 	      &84.67	&96.82	&92.83	&99.76	&61.08\\
       &   KNN-Class 	      &83.72	&96.83	&93.05	&99.77	&57.58\\
       &   SPB	              &81.28	&94.62	&91.48	&99.35	&64.38\\
       &   ODIN	              &36.67	&68.01	&67.26	&95.57	&23.04\\
       &   PCA                &89.94	&97.81	&94.52	&99.84	&70.83 \\
       &   Integrated(Our)         &\textbf{99.61}	&\textbf{98.47}	&\textbf{97.36}	&\textbf{99.89}	&\textbf{67.9}\\
       \\ \hline \\
LSUN &   Mahala-Tied    &78.38	&96.17	&91.98	&98.73	&85.11\\
       &   Mahala-Class	&81.51	&96.71	&92.44	&98.91	&87.63\\
       &   KNN-Tied     &77.61	&95.98	&91.34	&98.61	&85.56\\
       &   KNN-Class    &78.77	&96.05	&91.45	&98.62	&85.71\\
       &   SPB	        &74.29	&91.58	&88.96	&94.19	&83.95\\
       &   ODIN	        &35.92	&68.6	&66.75	&82.37	&53.12\\
       &   PCA          &82.93	&96.88	&92.74	&98.97	&88.27 \\
       &   Integrated(Our)           &\textbf{85.46}	&\textbf{97.09}	&\textbf{93.17}	&\textbf{99.03}	&\textbf{89.03}\\
          \\ \hline \\
CIFAR10   &   Mahala-Tied &85.03	&97.05	&93.15	&99.04	&88.62\\
       &   Mahala-Class	  &86.84	&97.41	&93.48	&99.15	&90.37\\
       &   KNN-Tied 	  &82.17	&96.65	&92.24	&98.87	&87.63\\
       &   KNN-Class 	  &83.24	&96.73	&92.38	&98.9	&87.67\\
       &   SPB	          &78.26	&92.92	&90.03	&95.06	&85.66\\
       &   ODIN	          &32.67	&66.75	&65.37	&80.69	&50.49\\
       &   PCA            &88.18	&97.55	&93.83	&99.2	&90.77\\
       &   Integrated(Our)           &\textbf{90.34}	&\textbf{97.64}	&\textbf{94.29}	&\textbf{99.17}	&\textbf{91.17}\\
       \\ \hline \\
\end{tabular}
\label{table:ablation-svhn}
\end{center}
\label{table:cifar10-densenet2}
\end{table}


\noop{
\newpage
\subsection{Intuition on the failure of different methods}
DKNN fig~\ref{fig:dknn_counter}
\begin{figure}[h]
\centering
\includegraphics[width=1\textwidth]{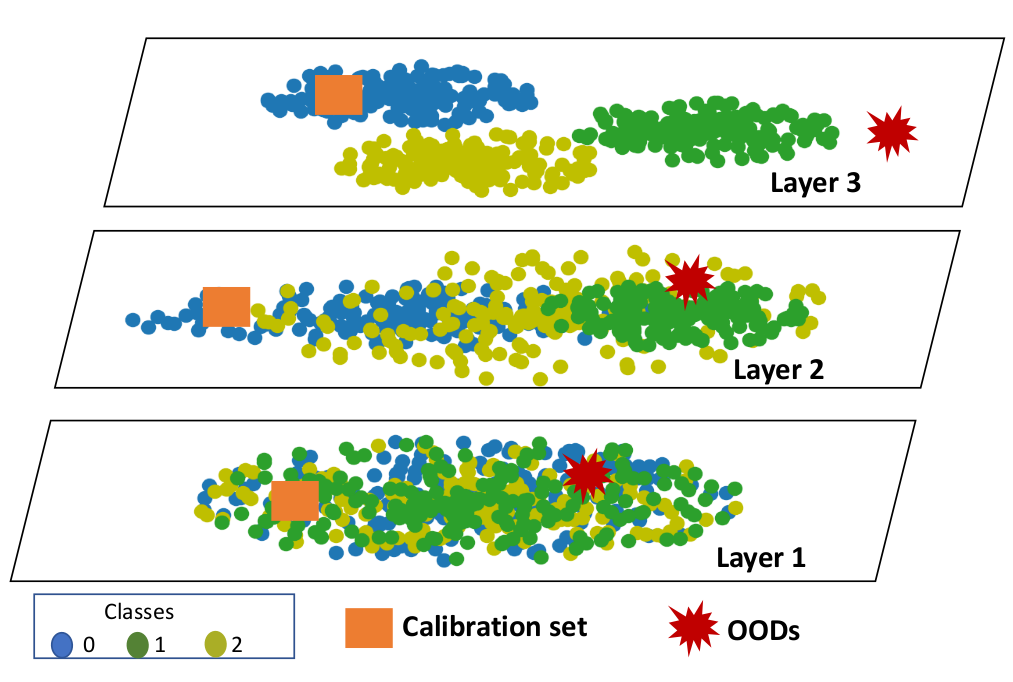}
\caption{OODs with the conformance measure in the labels of the nearest neighbors similar to the in-distribution samples, as visualized in 2D space}
\label{fig:dknn_counter}
\end{figure}

Mahalanobis fig~\ref{fig:mahalanobis_counter}, plot with the least probable OODs generated from regressor trained with Mahalanobis distance
\begin{figure}[h]
\centering
\includegraphics[width=1\textwidth]{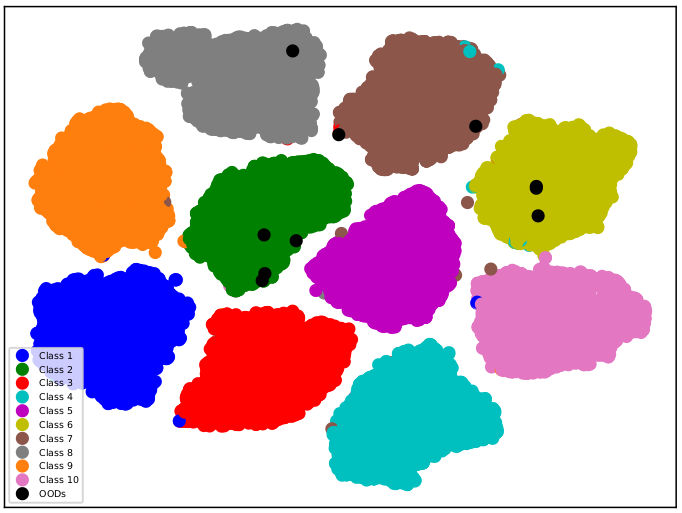}
\caption{OODs with mahalanobis distance of their features from the penultimate layer close to the corresponding mahalanobis distance of the in-distribution samples, as visualized in the 2D space}
\label{fig:mahalanobis_counter}
\end{figure}

Calibration by temp scaling/ODIN/Baseline/Softmax fig~\ref{fig:baseline_counter}, generated with the spiral data and a classifier trained on this data
\begin{figure}[h]
\centering
\includegraphics[width=1\textwidth]{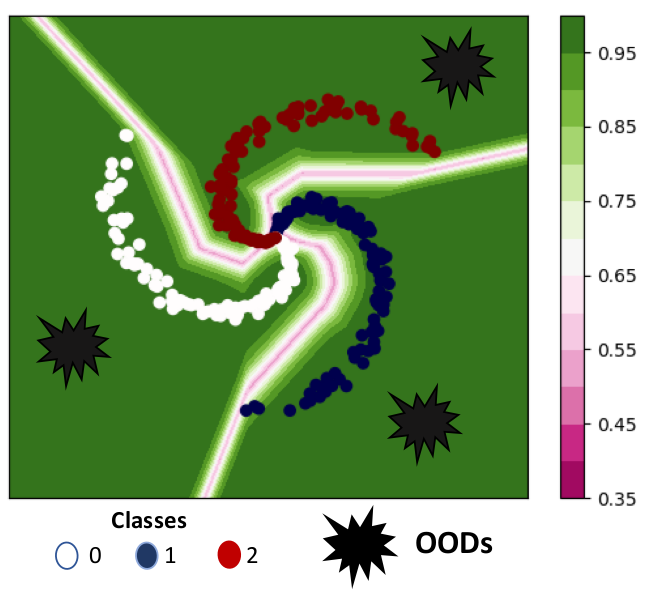}
\caption{OODs lying within the classification boundary of a particular class. Similar to the in-distribution samples, the softmax score of these OODs is very high ($\geq 0.95$).}
\label{fig:baseline_counter}
\end{figure}

LID fig~\ref{fig:lid_counter}
\begin{figure}[h]
\centering
\includegraphics[width=1\textwidth]{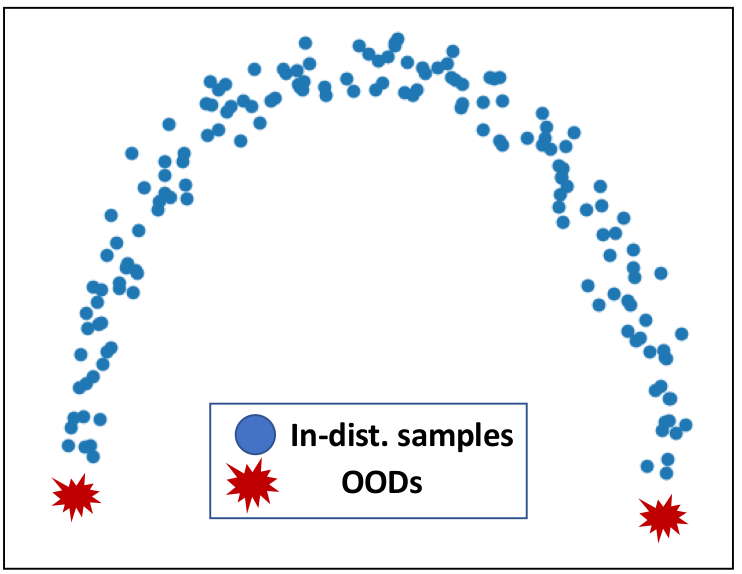}
\caption{OODs with local intrinsic dimensionality measure close to the in-distribution samples}
\label{fig:lid_counter}
\end{figure}

PCA fig~\ref{fig:pca_counter}
\begin{figure}[h]
\centering
\includegraphics[width=1\textwidth]{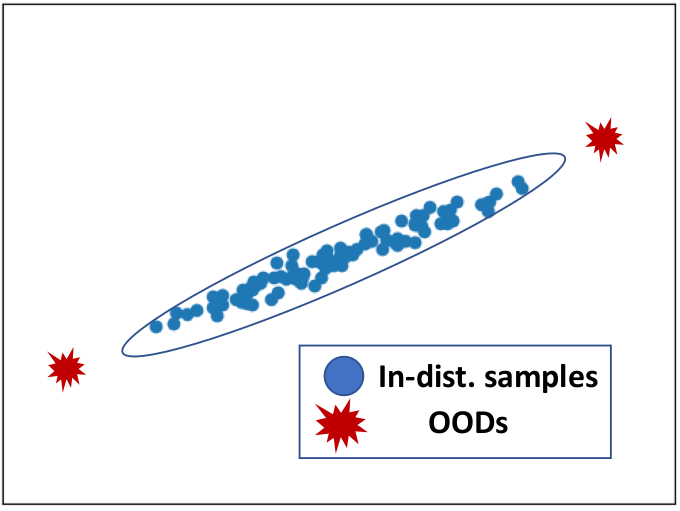}
\caption{OODs lying along the principal component of the in-distribution samples}
\label{fig:pca_counter}
\end{figure}
}

\end{document}